\documentclass[a4paper,10pt]{fullarticle}
\usepackage[british]{babel}
\usepackage{csquotes}
\usepackage{sciencestuff}

\definecolor{dartmouthgreen}{rgb}{0.05, 0.5, 0.06}
\definecolor{cobalt}{rgb}{0.0, 0.28, 0.67}
\definecolor{dandelion}{rgb}{0.94, 0.88, 0.19}
\definecolor{cinnamon}{rgb}{0.82, 0.41, 0.12}
\definecolor{ultramarineblue}{rgb}{0.25, 0.4, 0.96}

\input{arxiv.def}
\hypersetup{%
  pdftitle={Coverage-oriented Uncertainty Quantification for Machine Learning: an Application to Critical Heat Flux},
  pdfkeywords={uncertainty quantification, representation learning, machine learning, scientific machine learning, critical heat flux},
  pdfauthor={Michele Cazzola, Alberto Ghione, Lucia Sargentini, Julien Nespoulous, Riccardo Finotello},
  pdfsubject={Uncertainty quantification},
}

\DeclareRobustCommand{\etal}{\emph{et~al.}\xspace}      
\DeclareRobustCommand{\bepu}{BEPU\xspace}               
\DeclareRobustCommand{\auce}{AUCE\xspace}               
\DeclareRobustCommand{\bhr}{BHR\xspace}                 
\DeclareRobustCommand{\bnn}{BNN\xspace}                 
\DeclareRobustCommand{\bp}{BP\xspace}                   
\DeclareRobustCommand{\chf}{CHF\xspace}                 
\DeclareRobustCommand{\clb}{CLB\xspace}                 
\DeclareRobustCommand{\cpred}{CP\xspace}                
\DeclareRobustCommand{\cvae}{CVAE\xspace}               
\DeclareRobustCommand{\dcal}{\mathcal{D}_{\text{cal}}}  
\DeclareRobustCommand{\dnb}{DNB\xspace}                 
\DeclareRobustCommand{\dout}{\emph{dryout}\xspace}      
\DeclareRobustCommand{\dptr}{\mathcal{D}_{\text{pre}}}  
\DeclareRobustCommand{\dtr}{\mathcal{D}_{\text{tr}}}    
\DeclareRobustCommand{\elbo}{ELBO\xspace}               
\DeclareRobustCommand{\ghr}{GHR\xspace}                 
\DeclareRobustCommand{\gpr}{GPR\xspace}                 
\DeclareRobustCommand{\hr}{HR\xspace}                   
\DeclareRobustCommand{\iid}{i.i.d.\xspace}              
\DeclareRobustCommand{\inform}{INF\xspace}              
\DeclareRobustCommand{\kl}{KL\xspace}                   
\DeclareRobustCommand{\lut}{LUT\xspace}                 
\DeclareRobustCommand{\mape}{MAPE\xspace}               
\DeclareRobustCommand{\mlp}{MLP\xspace}                 
\DeclareRobustCommand{\ml}{ML\xspace}                   
\DeclareRobustCommand{\mpiw}{MPIW\xspace}               
\DeclareRobustCommand{\mse}{MSE\xspace}                 
\DeclareRobustCommand{\mtl}{MTL\xspace}                 
\DeclareRobustCommand{\nll}{NLL\xspace}                 
\DeclareRobustCommand{\nn}{NN\xspace}                   
\DeclareRobustCommand{\nrc}{NRC\xspace}                 
\DeclareRobustCommand{\picp}{PICP\xspace}               
\DeclareRobustCommand{\pinn}{PINN\xspace}               
\DeclareRobustCommand{\pint}{PI\xspace}                 
\DeclareRobustCommand{\qd}{QD\xspace}                   
\DeclareRobustCommand{\relu}{ReLU\xspace}               
\DeclareRobustCommand{\resnet}{ResNet\xspace}           
\DeclareRobustCommand{\rmse}{RMSE\xspace}               
\DeclareRobustCommand{\rmspe}{RMSPE\xspace}             
\DeclareRobustCommand{\svr}{SVR\xspace}                 
\DeclareRobustCommand{\tl}{TL\xspace}                   
\DeclareRobustCommand{\uqf}{UQF\xspace}                 
\DeclareRobustCommand{\uq}{UQ\xspace}                   

\DeclareRobustCommand{\cda}[1]{\hat{C}_{#1}^{\qty(\alpha)}}            
\DeclareRobustCommand{\cdas}[1]{\hat{C}_{#1}^{\qty(\alpha) \qty[\mathcal{S}]}}   
\DeclareRobustCommand{\qda}[1]{\hat{q}_{#1}^{\qty(1-\alpha)}}          
\DeclareRobustCommand{\qdas}[1]{\hat{q}_{#1}^{\qty(1-\alpha) \qty[\mathcal{S}]}} 

\DeclareMathOperator{\quant}{Q}

\title{%
  Learning Complex Physical Regimes via Coverage-oriented Uncertainty Quantification \\ \relax
  {\Large An application to the Critical Heat Flux}
}

\author[1]{Michele Cazzola\emailfoot{michele.cazzola@cea.fr}}
\author[1]{Alberto Ghione\emailfoot{alberto.ghione@cea.fr}}
\author[1]{Lucia Sargentini\emailfoot{lucia.sargentini@cea.fr}}
\author[2]{Julien Nespoulous\emailfoot{julien.nespoulous@cea.fr}}
\author[2]{Riccardo Finotello\emailfoot{riccardo.finotello@cea.fr}}
\affil[1]{%
Universit\'{e} Paris Saclay, \textsc{Cea},
\protect \\
Service Thermo-hydraulique et M\'{e}canique des Fluides (\textsc{Stmf}),
\protect \\
Gif-sur-Yvette, F-91191, France
}
\affil[2]{%
Universit\'{e} Paris Saclay, \textsc{Cea},
\protect \\
Service de G\'{e}nie Logiciel pour la Simulation (\textsc{Sgls}),
\protect \\
Gif-sur-Yvette, F-91191, France
}

\date{}

\begin{document}

\newgeometry{top=2cm,bottom=3cm}

\maketitle

\begin{abstract}

    A central challenge in scientific machine learning (\ml) is the correct representation of physical systems governed by complex, multi-regime behaviours.
    In these scenarios, standard data analysis techniques often fail to capture the changing nature of the data, as the system's response varies significantly across the state space due to the inherent stochasticity and the different physical regimes.
    Uncertainty quantification (\uq) should thus not be viewed merely as a safety assessment, but as a fundamental support to the learning task itself, guiding the model to internalise the underlying behaviour of the data.
    We address this challenge by focusing on the \emph{Critical Heat Flux} (\chf) benchmark and dataset presented by the OECD/NEA \emph{Expert Group on Reactor Systems Multi-Physics}.
    This case study represents a rigorous test for scientific \ml due to the highly non-linear dependence of \chf on the inputs and the existence of distinct microscopic physical regimes.
    These regimes exhibit diverse statistical profiles, a complexity that requires advanced \uq techniques to internalise the underlying data behaviour and ensure reliable predictions.

    In this work, we conduct a comparative analysis of \uq methodologies to determine their impact on physical representation.
    We contrast post-hoc methods, specifically conformal prediction, against end-to-end coverage-oriented pipelines, including (Bayesian) heteroscedastic regression and quality-driven loss functions.
    These end-to-end approaches treat uncertainty not as a final metric, but as an active component of the optimisation process, modelling the prediction and its statistical behaviour simultaneously.

    We show that while post-hoc methods ensure rigorous statistical calibration, coverage-oriented learning effectively reshapes the model's representation to match the complex physical regimes.
    The result is a model that delivers not only high predictive accuracy but also a physically consistent uncertainty estimation that adapts dynamically to the intrinsic variability of the \chf phenomenon.
\end{abstract}


\highlights{We show that uncertainty quantification can be used actively during the learning phase to model complex physical regimes. We consider the computation of the critical heat flux as a use case, and we show that coverage-oriented uncertainty quantification can be used to make the model more accurate and reliable.}

\clearpage

\restoregeometry

\tableofcontents

\clearpage

\section{Introduction}\label{sec:intro}

Machine learning (\ml) has deeply reshaped the way scientific data is analysed, providing powerful and flexible tools to model complex dependencies hidden in large and high-dimensional experimental datasets.
Unlike traditional empirical correlations and phenomenological models, often constrained by rigid or overly simplistic functional forms, data-driven approaches can adaptively model the complex behaviours typical of physical phenomena.
This flexibility has driven their widespread adoption across scientific domains, from particle physics to nuclear engineering, where they serve not only as surrogates but also as tools for hypothesis generation and experimental design.
However, the known black-box nature of many \ml methods has raised concerns about their reliability and interpretability, especially as they transition from auxiliary tools to core components of scientific workflows.

Scientific \ml is increasingly tasked with constructing surrogate models that do not merely interpolate experimental data, but capture the underlying physical laws and statistical behaviours of the system~\cite{uq-survey}.
While standard \ml regression approaches prioritise the minimisation of a global error metric, such as the mean squared error, this objective can be insufficient for physical applications characterised by complex, multi-regime dynamics.
In these scenarios, the observed data often exhibits varying degrees of stochasticity depending on the region of the input state space.
Consequently, a reductive approach that aims solely to minimise prediction intervals risks masking the intrinsic variability of the phenomenon.
By treating natural fluctuations as noise to be suppressed, a model may achieve high nominal precision but fails to learn a correct representation of the physical reality, leading to overconfident predictions in turbulent or unstable regimes~\cite{bishop-ml}.
Therefore, a robust scientific model must internalise this variability, treating the uncertainty not as a residual error, but as a fundamental component of the physical representation.

In this framework, uncertainty quantification (\uq) serves as a critical tool for assessing the trustworthiness and physical consistency of the learned representation~\cite{uq-survey}.
Rather than being a mere post-processing step for safety compliance, \uq provides a quantitative measure of how well the model's confidence aligns with the observed data distribution and a way to include this uncertainty directly into the learning process.
A rigorous criterion for this assessment is statistical coverage, defined as the probability that the true value of a physical quantity falls within the predicted uncertainty bounds.
Ensuring consistent coverage across the entire input domain is essential for establishing trust in the model, particularly in high-stakes scientific applications.
From a physical perspective, coverage-based methods allow the model to reveal changes in system behaviour: a localised expansion of uncertainty bounds is often a direct signature of a transition to a different regime, more stochastic or out-of-distribution with respect to the training data, where the model loses confidence in its own predictions.
One possible example of these changes may be the onset of physical instabilities or phase transitions.
Thus, achieving correct coverage is not just a statistical requirement but a means to validate that the model has correctly identified and internalised the distinct physical regimes present in the data.

We argue that achieving this rigorous coverage can be approached in two distinct ways, with different impacts on the model's representation power.
The first is a \emph{post-hoc} approach, such as conformal prediction (\cpred)~\cite{angelopoulos2022gentleintroductionconformalprediction}, which calibrates the uncertainty of a frozen, pre-trained model.
While invaluable for establishing statistical guarantees and ``trustworthiness'' on existing black boxes, this method does not alter the model's internal understanding of the physics.
It treats the model as a static entity, applying corrections to its outputs without influencing its learned features or representations.
Thus, the model remains oblivious to the underlying multi-regime nature of the data, though the uncertainty bounds are an attempt to reflect changes in the data behaviour and model trustworthiness.
The second, and more transformative approach, is to actively learn the uncertainty during training.
By integrating coverage-oriented objectives directly into the loss function, for instance through quality-driven techniques (\qd)~\cite{qd-loss} or (Bayesian) heteroscedastic regression (\hr)~\cite{heteroscedastic-regression}, we can force the model to acknowledge and adapt to the heteroscedastic nature of the data.
This active learning process does more than just quantify error; it improves the representation power of the model, allowing it to adapt its internal features to the changes in the behaviour of the system across the input space~\cite{bengioreplearn}, effectively ``learning'' the phase changes of the physics for the downstream task.

We test these hypotheses on the prediction of the \emph{Critical Heat Flux} (\chf), a phenomenon that serves as a rigorous benchmark for multi-regime scientific \ml.
\chf is characterised by a highly non-linear dependence on its inputs and arises from distinct microscopic mechanisms, \emph{departure from nucleate boiling} (\dnb) and \dout, which exhibit vastly different statistical profiles~\cite{collier, todreas}.
Recognising this complexity, the OECD/NEA \emph{Expert Group on Reactor Systems Multi-Physics} established a benchmark to develop \ml strategies capable of navigating these regimes~\cite{benchmark}.
In the specific context of nuclear safety, the alignment between predicted uncertainty and empirical variability is formalised in the \emph{Best Estimate plus Uncertainty} (\bepu) methodology~\cite{d2008best, sapium}.
Using this challenging dataset~\cite{groeneveld_nrc}, we demonstrate that while \cpred acts as a post-hoc implementation of the \bepu framework ensuring safety, coverage-oriented learning yields models that are physically consistent and robust by design, offering a representation-enhanced approach to the same rigorous safety standards.
This article also represents the first case study of the application of \qd loss to the \chf regression problem, showing how this method can be used to achieve coverage-based learning and to improve the physical representation of the model.

The article is organised as follows:
\begin{itemize}
    \item Section~\ref{sec:related-work} briefly recalls the main research works conducted on \ml and \uq for the prediction of the \chf in thermal hydraulics;
    \item Section~\ref{sec:dataset} details the characteristics of the OECD/NEA benchmark dataset and the preprocessing required to expose its physical structure;
    \item Section~\ref{sec:method} establishes a theoretical overview of the \ml and \uq techniques used in the article. This includes our proposal of an approach based on coverage and quality-driven methods to correctly assess the uncertainties for any given model;
    \item Section~\ref{sec:eval-metrics} defines the evaluation metrics used to assess both predictive performance and uncertainty calibration;
    \item Section~\ref{sec:results} presents the comparative analysis, evaluating both predictive performance and physical consistency;
    \item Section~\ref{sec:conclusion} summarises the impact of coverage-based learning on scientific representation.
\end{itemize}

\section{Related Work}\label{sec:related-work}

Traditionally, the prediction of \chf has relied on phenomenological approaches, which are often constrained to specific operating conditions.
Early tools were models based on empirical correlations and simplified analytical models that captured certain aspects of the underlying physics, with each model having its own limited validity range.
For instance, some of the pioneering correlations for \chf prediction, such as the \emph{Biasi correlation}~\cite{biasi}, were based on empirical models with minimal validity domains.
These were succeeded by more comprehensive look-up tables (\lut), such as the \emph{2006 Groeneveld \lut}~\cite{groeneveld}, which fit polynomial functions to larger databases.
While these methods expanded the domain of application and are still widely used in \chf regression, they remain static representations that struggle to capture the complex, multi-regime nature of the phenomenon: on the benchmark dataset (see Section~\ref{sec:dataset}), they reach 66\% (Biasi) and 36\% (Groeneveld \lut) of root mean squared percentage error (\rmspe, see~\eqref{eq:metric-rmspe} for the definition).

The inherent limitations of these fixed correlations motivated the exploration of data-driven strategies.
Early \ml works investigated hybrid approaches combining data with domain-specific knowledge~\cite{he_lee, zhao_informed, park, kim, zubair}.
These studies demonstrated the potential of \ml to outperform traditional models by identifying effective architectures and learning strategies for the \chf regression task.
However, these works relied on \emph{ad-hoc} datasets compiled by the respective authors from different, publicly available sources.
They thus lacked the rigorous comparability required for a standardised benchmark.

The release of the U.S.\ Nuclear Regulatory Commission (\nrc) dataset~\cite{groeneveld_nrc} established a common ground for evaluation.
Grosfilley~\etal~\cite{grosfilley-lecorre} developed and compared various \ml methods, including support vector regression (\svr), Gaussian process regression (\gpr), and neural networks (\nn), achieving a significant error reduction (\rmspe of 12\%) compared to the Groeneveld \lut.
Cassetta~\etal~\cite{cassetta} further advanced the state-of-the-art by employing deep learning models with residual connections (\resnet) and physics-informed frameworks (\pinn)~\cite{zhao_informed}.
While these works solidified the predictive capability of \ml, their primary focus remained on minimising global error metrics, treating the underlying physical regimes as a single optimisation problem.

Thanks to the availability of standardised data, recent research has also begun to address the reliability of these predictions through \uq.
Ahmed~\etal~\cite{ahmed-deep-ensemble} utilised a technique based on a specific initialisation of deep ensembles for robust model selection, while Alsafadi~\etal~\cite{alsafadi-cvae} explored conditional variational autoencoders (\cvae)~\cite{cvae} to account for domain generalisation and stochasticity.
The results were compared to those obtained using a multi-layer perceptron (\mlp) architecture, that relies on a deep ensemble to estimate the uncertainty.
This work achieves similar results to the previous ones on the \chf prediction task.
Furlong~\etal~\cite{furlong-pinn-uq} focused on the specific regime of \dout points, retrieved with expert judgement from the \nrc dataset, comparing deep ensembles and Bayesian \nn (\bnn) to separate aleatoric and epistemic uncertainty components, and focus on the latter.
However, these approaches predominantly focus on disentangling aleatoric and epistemic uncertainty components, and usually focussing only on the latter, to assess model reliability, rather than using them as a mechanism to enforce physical consistency in the learning process.

Although current data-driven approaches have significantly improved \chf prediction accuracy, standard \uq applications often prioritise the minimisation of the epistemic uncertainty.
This strategy risks masking the true aleatoric uncertainty inherent to the different physical regimes, potentially leading to overconfident predictions in complex transition zones.
To address this, we take inspiration from the emerging paradigm in scientific \ml that seeks rigorous statistical guarantees, such as the adoption of \cpred as a standard calibration layer in High-Energy Physics~\cite{cp_hep_2025}.
We extend this perspective by integrating coverage-oriented objectives, specifically \qd learning and (Bayesian) \hr, directly into the optimisation objective.
Rather than treating uncertainty as a post-hoc safety margin, this approach forces the model to internalise the stochastic profile of the data, learning a representation that is both statistically valid and physically consistent with the multi-regime nature of the \chf.

\section{The U.S.\ Nuclear Regulatory Commission Dataset}\label{sec:dataset}

\begin{figure}[t]
    \centering
    \begin{tikzpicture}
        \fill[very thick, red] (0.15, 0) rectangle (0.0, 5) node[midway, above, rotate=90] {\Large \textsc{Heat Flux}};
        \foreach \i in {0.5, 1.5, 2.5, 3.5, 4.5} {
                \draw[draw=cinnamon, fill=dandelion, very thick] (0.2, \i) -- ++(0, -0.15) -- ++(1.5, 0) -- ++(0, -0.1) -- ++(0.5, 0.25) -- ++(-0.5, +0.25) -- ++(0, -0.1) -- ++(-1.5, 0) -- cycle;
            }

        \fill[very thick, black!50] (2.6, 0) rectangle (2.8, 5);
        \fill[very thick, black!50] (4.6, 0) rectangle (4.8, 5);
        \fill[ultramarineblue!35] (2.8, 0) rectangle (4.6, 5) node[midway, above, black, rotate=90] {\Large \textsc{Coolant Liquid}} node[midway, below, black, rotate=90] {\Large $\longrightarrow \quad \longrightarrow \quad \longrightarrow$};

        \draw[thick, |<->|] (5.0, 0) -- (5.0, 5) node[midway, above right, black] (heated length) {\textsc{Heated Length}};
        \node[below=0.15em of heated length, black] {\Large $L$};

        \draw[thick, |<->|] (2.8, -0.2) -- (4.6, -0.2) node[midway, below, black] (tube diameter) {\textsc{Tube Diameter}};
        \node[below=0.15em of tube diameter, black] {\Large $D$};

        \draw[ultramarineblue, ultra thick, ->, rounded corners=3mm] (6.0, -1.75) node[right, anchor=south west] {\textsc{Mass Flow Rate}~ {\Large $G$}} node[right, anchor=north west] {\textsc{Inlet Temperature}~ {\Large $T_{\text{in}}$}} -- (4.3, -1.75) -- (4.3, -0.9);

        \draw[ultramarineblue, ultra thick, ->, rounded corners=3mm] (4.3, 5.2) -- (4.3, 6.05) -- (6.0, 6.05) node[right, anchor=south west] {\textsc{Pressure}~ {\Large $P$}} node[right, anchor=north west] {\textsc{Outlet Vapour Quality}~ {\Large $\mathfrak{X}$}};
    \end{tikzpicture}
    \caption{Experimental setting of \chf measurements in the \nrc dataset.}\label{fig:chf-exp}
\end{figure}
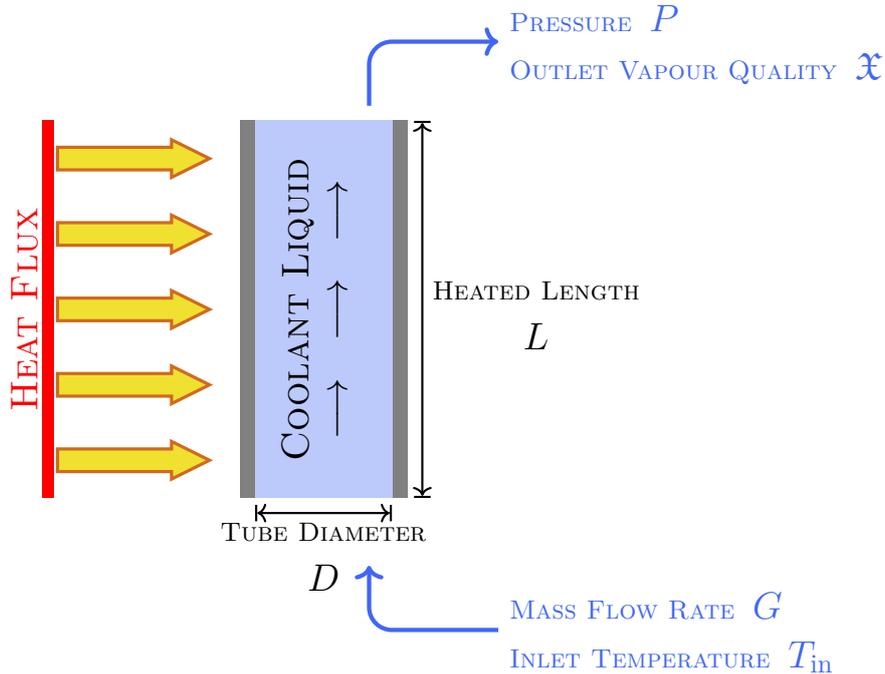

The data used in this work has been published by the U.S.\ \nrc, and it will be referred to as \emph{\nrc dataset}~\cite{groeneveld_nrc}.
Entries come from 60 different sources: each one represents an experiment carried out in the uniformly heated linear tube setting shown in Figure~\ref{fig:chf-exp} and discussed in Section~\ref{sec:intro}, during the past 70 years.
For each source, a highly variable number of samples (trials) is provided.
The dataset contains a total of \num{24579} entries.
A preliminary check for outliers, missing, incomplete or duplicate values has already been performed by the benchmark organisers~\cite{benchmark}.

The dataset provides the following input features (with the corresponding units of measurement):

\begin{itemize}
    \item $D$ [\si{\meter}]: diameter of the tube;
    \item $L$ [\si{\meter}]: heated length of the tube;
    \item $P$ [\si{\kilo\pascal}]: pressure of the coolant inside the tube;
    \item $G$ [\si{\kilo\gram\per\meter\squared\per\second}]: mass flux of the coolant, that is the mass flow rate per area and per time unit;
    \item $T_{\text{in}}$ [\si{\celsius}]: inlet temperature of the coolant;
    \item $\mathfrak{X}$ [dimensionless quantity]: outlet thermodynamic quality;
    \item $\Delta h_{\text{sub,\,in}}$ [\si{\kilo\joule\per\kilo\gram}]: inlet subcooling.
\end{itemize}

Inputs thus divide into geometrical parameters ($D$, $L$) and physical properties ($P$, $G$, $T_{\text{in}}$).
To complement these parameters, additional thermodynamical quantities are computed to characterise the system fully.
In particular, the outlet vapour quality is a dimensionless quantity that represents the fraction of vapour in the system from an energetic point of view.
It is computed as:
\begin{equation}
    \label{eq:outlet-quality}
    \mathfrak{X}
    =
    \frac{h_{\text{out}} - h_{\text{sat}}^{(l)}\qty(P)}{h_{\text{sat}}^{(v)}\qty(P) - h_{\text{sat}}^{(l)}\qty(P)},
\end{equation}
where $h_{\text{out}}$ is the outlet specific enthalpy, $h_{\text{sat}}^{(l)}$ is the specific enthalpy of the liquid coolant at the saturation point, and $h_{\text{sat}}^{(v)}$ is the saturation specific enthalpy of the vapour.
Both $h_{\text{sat}}^{(l)}$ and $h_{\text{sat}}^{(v)}$ depend on the pressure $P$.
The outlet specific enthalpy $h_{\text{out}}$ is derived from the following equation:
\begin{equation}
    \label{eq:outlet-specific-enthalpy}
    h_{\text{out}} = h_{\text{in}} + \frac{4 \Phi_q L}{G D},
\end{equation}
which is obtained solving the heat balance equation for the uniformly heated tube for the power $\mathcal{Q}$:
\begin{equation}
    \label{eq:outlet-specific-enthalpy-balance}
    \mathcal{Q}
    =
    (h_{\text{out}} - h_{\text{in}}) \frac{\pi G D^2}{4}
    =
    \int \dd{S}\, \Phi_q,
\end{equation}
where $\Phi_q$ indicates the heat flux, $S$ the generic heated surface, and $h_{\text{in}}$ the inlet specific enthalpy.
When negative, vapour quality indicates the distance of the liquid from its saturation point.
Another relevant quantity, the inlet subcooling also measures the separation of the coolant entering the heated tube from its saturation point from an energetic point of view.
It is expressed as follows:
\begin{equation}
    \label{eq:delta-h-sub}
    \Delta h_{\text{sub,\,in}} = h_{\text{sat}}^{(l)}(P) - h_{\text{in}},
\end{equation}
where $h_{\text{in}}$ is the inlet specific enthalpy.
The output feature, target of the regression, is the numerical value of the \chf [\si{\kilo\watt\per\meter\squared}].

Particular attention must be paid when selecting the input features for the \ml models.
The outlet quality $\mathfrak{X}$ is computed through~\eqref{eq:outlet-quality} and~\eqref{eq:outlet-specific-enthalpy}.
At its critical point, the equation includes both the \chf and the input features: it is important to exclude the features used for computing $\mathfrak{X}$ (if used as input parameter of the model), otherwise the \chf could be easily estimated~\cite{benchmark}.
For this reason, the inlet properties ($T_{\text{in}}$ and $\Delta h_{\text{sub,\,in}}$) will not be used as input parameters of the models.

A detailed and exhaustive analysis of the dataset features has been previously conducted, noticing a relevant inverse linear correlation between \chf and both outlet quality and heated length~\cite{cassetta}.
A filtered version of the \nrc dataset is utilised in this work.
This filtering process is essential for restoring physical consistency to the training domain, as entries with negative subcooling or from unreliable sources often exhibit flow instabilities that mask the intrinsic physical mechanisms of the \chf phenomenon.
The dataset is thus generated by removing two groups of samples from the original \nrc database:

\begin{itemize}
    \item those related to a value of $\Delta h_{\text{sub,\,in}} \le 0$: flow instability and deviations from the actual \chf behaviour may be found in this regime~\cite{boure-flow-instability},
    \item those collected from the following sources: Alessandrini \etal (1963), Soderquist (1994), and Kureta (1997)~\cite{benchmark}: these sources cannot be considered reliable for our task because of imprecisions in the data collection process (e.g.\ numeric values directly retrieved from scatter plots), negative inlet subcooling, or the use of tubes that are too short and thus susceptible to flow instability~\cite{groeneveld_nrc}.
\end{itemize}

After excluding these samples, the filtered dataset contains \num{22872} points in total.

\section{Methodology}\label{sec:method}

Models developed in this article deal with the tabular data in the \nrc dataset.
This section focuses on the theoretical grounds of the employed \ml and \uq methods.
In particular, we shall focus on defining the concepts of residual networks (\resnet) for tabular data, conformal prediction (\cpred), heteroscedastic regression (\hr), quality-driven prediction and uncertainty estimation (\qd), and Bayesian heteroscedastic regression (\bhr).

In what follows, we shall refer to scalar quantities using lowercase letters (e.g.\ $z$), vectors with lowercase bold letters (e.g.\ $\vb{z}$), and matrices with uppercase letters (e.g.\ $Z$).
This will allow us to distinguish between the different types of realisations of the random variables (e.g.\ $\mathcal{Z}$).
Unless otherwise specified, data samples are vector quantities $\vb{x} \in \mathbb{R}^d$ organised in the rows of a matrix $X \in \mathbb{R}^{N \times d}$, where $N$ is the number of samples and $d$ the number of features:
\begin{equation}
    \label{eq:data-matrix}
    X = \mqty(%
    \vb{x}_1 \\
    \vb{x}_2 \\
    \vdots \\
    \vb{x}_N
    ) = \mqty(%
    x_{11} & x_{12} & \cdots & x_{1d} \\
    x_{21} & x_{22} & \cdots & x_{2d} \\
    \vdots & \vdots & \ddots & \vdots \\
    x_{N1} & x_{N2} & \cdots & x_{Nd}
    ) \in \mathbb{R}^{N \times d}.
\end{equation}
Moreover, we shall use the notation $\hat{\omega}$ to indicate the estimated (optimal) value of the quantity $\omega$.
This could, for instance, be the optimal parameters $\hat{\theta}$ of a trained \ml model, whose prediction will be denoted as $\mu(\vb{x}; \hat{\theta})$.

\subsection{Residual Networks for Physics Surrogate Models}\label{subsec:method-resnet}

For the analysis of the tabular data, we consider a set of architectures based on fully-connected \mlp~\cite{rosenblatt1958perceptron}.
This is a combination of linear affine transformations and activation functions, which ensure the non-linear behaviour of the network and are usually applied on the output of the linear layers.
A single layer of the network acts on an input vector $\vb{x}^{(\ell-1)} \in \mathbb{R}^{n^{(\ell-1)}}$ as:
\begin{equation}
    \label{eq:mlplayer}
    \vb{x}^{(\ell)} \equiv f_{\theta^{(\ell)}}^{(\ell)}\left(\vb{x}^{(\ell-1)}\right) = a^{(\ell)}\left( W^{(\ell)} \vb{x}^{(\ell-1)} + b^{(\ell)} \right) \in \mathbb{R}^{n^{(\ell)}},
\end{equation}
where $\ell = 1, 2, \ldots, L$ identifies the layer in the network, $a^{(\ell)}$ is the activation function applied entry-wise to the output of the affine transformation, and
\begin{equation}
    \theta^{(\ell)} = \left\lbrace W^{(\ell)} \in \mathbb{R}^{n^{(\ell)} \times n^{(\ell-1)}},\; b^{(\ell)} \in \mathbb{R}^{n^{(\ell)}} \right\rbrace
\end{equation}
are the \emph{weights} and \emph{bias} tensors of the $\ell$-th layer.
In this example, $\vb{x}^{(\ell-1)}$ is either the input data (if $\ell = 1$, with $d = n^{(0)}$ in~\eqref{eq:data-matrix}) or the output of the previous layer, in matrix form.
In this work, a \relu activation function~\cite{relu} is placed after each hidden layer.
The output layer $f_{\theta^{(L)}}^{(L)}$ can either be a linear layer, where the corresponding activation function $a^{(L)}$ is the identity function, or a particular activation function, used to constrain the output of the model to a specific range.
To constrain the model to learn to predict the \chf as a strictly positive quantity, a \emph{softplus} activation function~\cite{softplus} is used after the output layer.
Its expression is:
\begin{equation}
    \label{eq:softplus}
    \mathrm{softplus}\qty(z)
    =
    \frac{1}{\beta} \ln \left( 1 + e^{\beta z} \right),
\end{equation}
where $\beta$ represents the smoothness factor of the function.
By definition, this ensures that all numerical predictions of the network are strictly positive.

The model prediction is then obtained by composition of the output of the layers, as:
\begin{equation}
    \vb{x}^{(L)}
    =
    \boldsymbol{\mu}(\vb{x}^{(0)}; \theta)
    =
    f_{\theta^{(L)}}^{(L)} \circ f_{\theta^{(L-1)}}^{(L-1)} \circ \cdots \circ f_{\theta^{(1)}}^{(1)}\left( \vb{x}^{(0)} \right)
    \in \mathbb{R}^{n^{(L)}},
\end{equation}
where $\theta = \bigcup_{\ell=1}^{L} \theta^{(\ell)}$ is the set of all parameters of the network, and $\vb{x}^{(0)}$ is the input data.

Dealing with deep \nn models may lead to the phenomenon of vanishing gradients.
It consists of the presence of gradient values close to zero, which hampers the learning process~\cite{vanishing-gradients}.
Residual blocks have been proposed to mitigate the issue, by providing an alternative path for the gradient flow during backpropagation.
\nn architectures based on this mechanism are called \emph{residual networks} (\resnet)~\cite{resnet}.
Although they were introduced for computer vision tasks with convolutional layers, they can also be adapted to linear layers.
In this work, we follow the same approach as~\cite{cassetta} and implement the residual block with a single linear layer preceded by batch normalisation and the activation function, according to~\cite{resnet-mappings}:
\begin{equation}
    \label{eq:resnetlayer}
    \vb{x}^{(\ell)} = f_{\theta^{(\ell)}}^{(\ell)}\left(\vb{x}^{(\ell-1)}\right) + \vb{x}^{(\ell-1)}.
\end{equation}
This residual unit serves as the fundamental building block for the surrogate models explored in this study.
Overall, they show an increased stability, notwithstanding the underlying complexity of the data.

\subsection{Loss Functions and Optimisation}\label{subsec:method-loss}

Training a \nn eventually boils down to the optimisation of a learning objective through the minimisation of a loss function.
Its formulation depends on the specific task for which the model is designed.
Since, for the moment, we focus on a regression problem with scalar output, the simplest possible choice is the \emph{mean squared error} (\mse) loss, which computes the average of the squared errors between the model predictions $\mu(\vb{x}; \theta)$ and the ground truths $y$:
\begin{equation}
    \label{eq:mse-loss}
    {L}_{\text{\mse}} (\theta; X, \vb{y}) = \frac{1}{N} \sum_{i=1}^{N} {\left( y_{i} - \mu(\vb{x}_i; \theta) \right)} ^{2},
\end{equation}
where $N$ is the number of samples.
However, since we are interested in minimising the relative error of the predictions, we can also use functions such as the root mean squared percentage error (\rmspe).
As already seen in~\cite{cassetta}, we use the definition:
\begin{equation}
    \label{eq:rmspe-loss}
    {L}_{\text{\rmspe}} (\theta; X, \vb{y}) = \sqrt{ \frac{1}{N} \sum_{i=1}^{N} {\left( \frac{y_{i} - \mu(\vb{x}_i; \theta)}{y_{i}} \right)} ^{2} }.
\end{equation}

Model parameters are updated using SGD~\cite{sgd}, Adam~\cite{adam}, or AdamW~\cite{adamw}, with the specific optimiser selected through a hyperparameter search.
Weight decay regularisation is employed to mitigate overfitting and improve the generalisation capability of the models~\cite{wd-regularization}.
To accelerate convergence and improve the regularisation of the model we use \emph{batch normalisation}~\cite{ioffe-bn}.

\subsection{Uncertainty Quantification Techniques}\label{subsec:method-uq}

\uq estimates how likely certain outcomes are if some aspects of the system are not completely known.
In the context of scientific \ml for multi-regime systems, we view \uq not merely as a tool for error estimation, but as a mechanism to internalise the stochastic profile of the data directly into the model's latent representation.
Statistical uncertainty can come from different sources, which can be classified into:

\begin{itemize}
    \item \emph{aleatoric uncertainty}: also known as \emph{data noise variance}, it originates from stochasticity in the data acquisition process,
    \item \emph{epistemic uncertainty}: also called \emph{model} or \emph{systematic uncertainty}, it arises from uncertainty on the model parameters, lack of knowledge or during training or inference process, sub-optimal design choices, and out-of-distribution test samples.
\end{itemize}

Both these contributions constitute the \emph{prediction uncertainty}, which expresses what the combination of architecture and data cannot hope to model~\cite{uq-survey}.

In regression problems, we consider observations
\begin{equation}
    y = \mu\left( \vb{x} \right) + \varepsilon_{\text{data}},
\end{equation}
generated by some ``true'' process $\mu$ we aim at approximating.
Here, $\vb{x}$ indicates an input data sample, and $\varepsilon_{\text{data}} \sim \mathcal{N} \left( 0, \sigma_{\text{data}}^2 \right)$ the noise around collected data.
The ground truth (i.e.\ the observed value) can be estimated by a trained \ml model $\mu(\vb{x}; \hat{\theta})$ as
\begin{equation}
    \label{eq:ml-model}
    y = \mu(\vb{x}; \hat{\theta}) + \varepsilon_{\text{pred}},
\end{equation}
where $\varepsilon_{\text{pred}} \sim \mathcal{N} \left( 0, \sigma_{\text{pred}}^2 \right)$ represents the total prediction error.
The prediction variance may be decomposed into~\cite{bishop-ml, qd-loss}:
\begin{equation}
    \label{eq:eq-pred-uncertainty}
    \sigma_{\text{pred}}^2(\vb{x}; \hat{\theta})
    =
    \sigma_{\text{data}}^2(\vb{x}) + \sigma_{\text{model}}^2(\vb{x}; \hat{\theta}).
\end{equation}
It therefore depends on both the uncertainty components.
Since the true function $\mu$ is generally unknown, and we can only make high-level assumptions about the data noise $\varepsilon_{\text{data}}$ (i.e.\ its distribution), only prediction uncertainty is generally directly treatable.
Even though data and systematic uncertainties cannot be entirely disentangled, some models can, however, provide a good approximation of model and aleatoric uncertainty, by separately estimating the two components (see Section~\ref{subsec:method-bhr}).

A good measure of the uncertainty associated to a \ml model is to define the data-dependent prediction interval (\pint) $\cda{\mathcal{D}_{N}}\qty(\vb{x})$, defined on a dataset $\mathcal{D}_{N}$ which contains $N$ samples, with significance level $\alpha \in \left(0, 1\right)$.
It enables to guarantee the coverage on an unseen sample $\left( \vb{x}_{N+1}, y_{N+1} \right) \sim P_{\vb{X}Y}$ with probability $1 - \alpha$, when $\mathcal{D}_{N}$ is the \iid sample of $N$ points from the joint distribution $P_{\vb{X}Y}$ used to build the \pint:
\begin{equation}
    \label{eq:prediction-interval}
    P \left(%
    y_{N+1} \in \cda{\mathcal{D}_{N}}\qty(\vb{x}_{N+1})
    \right)
    \ge
    1 - \alpha.
\end{equation}
In a \ml context like this article, the dataset $\mathcal{D}_{N}$ contains (or coincides with) the training set of a \ml model~\cite{lei2017distributionfreepredictiveinferenceregression}.
A \pint quantifies the prediction uncertainty associated with a given model used to compute them.
For this reason, the whole uncertainty contribution must be captured to calculate a \pint, and \uq methods described in the next sections will provide different ways of computing such quantity.

\subsection{Heteroscedasticity tests}\label{subsec:method-ht}

Given the error $\varepsilon \sim \mathcal{N} \left( 0, \sigma^2 \right)$ previously defined, we identify two possible behaviours~\cite{homoscedasticity}:

\begin{itemize}
    \item \emph{homoscedasticity}: the standard deviation $\sigma$ is constant across observations, independent of the input features ($\nabla_{\vb{x}} \sigma = 0$);
    \item \emph{heteroscedasticity}: the standard deviation $\sigma$ depends on the input features, and we can express the error as $ \varepsilon \sim \mathcal{N} (0,\sigma^2 \left( \vb{x} \right)) $.
\end{itemize}

Knowing whether our models are homoscedastic is crucial to perform \uq.
For this reason, we use the \emph{Breusch-Pagan test} (\bp) to identify heteroscedastic models~\cite{breusch-pagan}.
This is a hypothesis test, whose null hypothesis is:
\begin{center}
    $\mathcal{H}_{0}:~$~\emph{``The model is homoscedastic''},
\end{center}
which corresponds to parametrise the variance of the residuals as $\operatorname{Var}\qty(\varepsilon) = \sigma^2 = \bar{\sigma}^2\, h(\vb{x})$, where $\bar{\sigma}$ is constant, and $h(\vb{x}) \equiv 1$ if the null hypothesis holds.

From the technical point of view, the test consists of fitting a linear regression model on the squared residuals $\qty{ \varepsilon_i^2 }_{i = 1, 2, \dots, N}$ of the original \ml model~\eqref{eq:ml-model}, using the same input features $\vb{x}$:
\begin{equation}
    \label{eq:bp-test-lm}
    \varepsilon_i^2
    =
    \beta_0 + \sum_{j=1}^{p} \beta_j x_{ij} + \xi_i,
\end{equation}
where $p$ is the number of input features (variables) in the model.
Formally, the null hypothesis holds if all regression coefficients $\beta_{j \ge 1}$ vanish.
The \bp test statistic (in its robust form) is then proportional to the \emph{coefficient of determination} $R^2$ of the linear model~\eqref{eq:bp-test-lm}~\cite{koenker1981bp}:
\begin{equation}
    \label{eq:bp-test-statistic}
    \operatorname{BP}
    =
    N R^2
    =
    N\,
    \sum\limits_{i = 1}^{N}
    \frac{{\qty(\varepsilon_i^2(\vb{x}; \hat{\beta}) - \sigma^2)}^2}{{\qty(\varepsilon_i^2 - \sigma^2)}^2},
\end{equation}
where $\varepsilon_i(\vb{x}; \hat{\beta})$ is the estimated value of $\varepsilon_i$ through the model~\eqref{eq:bp-test-lm}.
It can be shown that, under the null hypothesis of homoscedasticity, the \bp statistic asymptotically follows a $\chi^2$ distribution with $p$ degrees of freedom~\cite{breusch-pagan,koenker1981bp}.
The test can thus be evaluated at a significance level $\alpha \in \left(0, 1\right)$ to determine whether we can reject the null hypothesis and, thus, consider the \ml models as heteroscedastic with a degree of confidence $1 - \alpha$.

\subsection{Conformal Prediction}\label{subsec:method-cp}

\cpred is a statistical learning method that enables the estimation of the uncertainty bounds of numerical models, within an arbitrary level of confidence.
The technique follows a non-parametric, distribution-free approach.
The variant of \cpred used in this work is generally known as \emph{split conformal prediction}.
We refer to it as \cpred, for simplicity, and it will serve here as our post-hoc statistical benchmark, enabling the calibration of pre-trained models without altering their internal representation.
In this realisation, the training set $\dtr$ is divided into a \emph{pretraining} set $\dptr$ used to train the \ml model, and a \emph{calibration} set $\dcal$ used to estimate the empirical quantiles~\cite{angelopoulos2022gentleintroductionconformalprediction}:
\begin{equation}
    \dtr = \dptr \cup \dcal, \quad \dptr \cap \dcal = \emptyset.
\end{equation}

The vanilla version of \cpred creates constant bounds, by definition of the \pint.
Let $\mathcal{R}_{\dcal}$ be the set of absolute residuals between the predictions of the \ml model $\mu(\vb{x}; \hat{\theta}_{\text{pre}})$, trained on $\dptr$ and thus parametrised by $\hat{\theta}_{\text{pre}}$, and the corresponding ground truths $y$ on $\dcal$:
\begin{equation}
    \mathcal{R}_{\dcal}
    =
    \left\lbrace
    \abs{y - \mu(\vb{x}; \hat{\theta}_{\text{pre}})},~
    \qty(\vb{x}, y) \in \dcal
    \right\rbrace.
\end{equation}
The width of the intervals is related to the empirical estimation of the quantiles $\quant$ of $\mathcal{R}_{\dcal}$.
This is expressed as:
\begin{equation}
    \label{eq:cp-quantile}
    \qda{\dcal}
    =
    \hat{\quant}_{(1-\alpha)\frac{m+1}{m}} \qty(\mathcal{R}_{\dcal})
    =
    \rank_{(1-\alpha)\frac{m+1}{m}}\qty( \mathcal{R}_{\dcal} ),
\end{equation}
where $1 - \alpha$ is the confidence level, and $m$ is the number of calibration samples.
The \pint on the training set can be computed as:
\begin{equation}
    \label{eq:cp-prediction}
    \cda{\dtr}(\vb{x})
    =
    \left[
        \mu(\vb{x}; \hat{\theta}_{\text{pre}}) - \qda{\dcal},~
        \mu(\vb{x}; \hat{\theta}_{\text{pre}}) + \qda{\dcal}
        \right].
\end{equation}

\begin{figure*}
    \centering
    \includegraphics[width=0.8\linewidth]{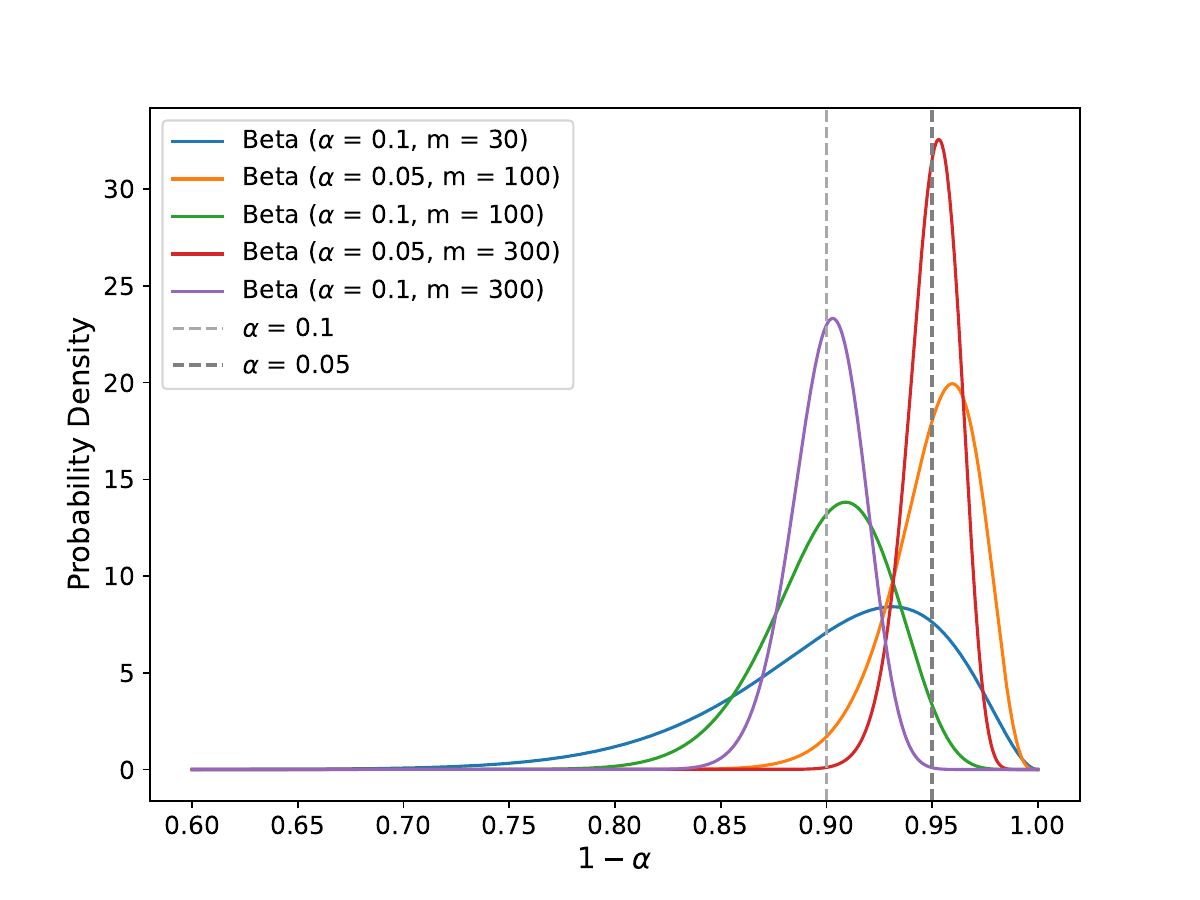}
    \caption{Beta probability density functions, as functions of the variables $\alpha$ and $m$.}\label{fig:beta-distribution}
\end{figure*}

\cpred provides guarantees about the coverage of the \pint with respect to an unseen sample.
In particular, given a generic, unseen data point $(\vb{x}_{N+1}, y_{N+1}) \not\in \dtr$, it holds that:
\begin{equation}
    \label{eq:cp-beta-coverage}
    P\qty(y_{N+1} \in \cda{\dtr}(\vb{x}_{N+1}) \mid \dtr)
    \sim
    \operatorname{Beta}\qty(l, N + 1 - l),
\end{equation}
where $l = \lceil (1 + N)(1 - \alpha) \rceil$.
Some examples of such distribution are shown in Figure~\ref{fig:beta-distribution}~\cite{angelopoulos2022gentleintroductionconformalprediction}.

This technique cannot account for variable uncertainty bounds, making it impossible to discriminate between regions where the model is more confident, or where more data are provided.
In a heteroscedastic setting, it is possible to employ a score function in the \cpred framework that accounts for the variability of the uncertainty:
\begin{equation}
    \label{eq:cp-score-function}
    S(\vb{x}, y)
    =
    \frac{| y - \mu(\vb{x}; \hat{\theta}_{\text{pre}})|}{\sigma(\vb{x}; \hat{\theta}_{\text{pre}})},
\end{equation}
where $| y - \mu(\vb{x}; \hat{\theta}_{\text{pre}})|$ is the absolute residuals previously defined, and $\sigma(\vb{x}; \hat{\theta}_{\text{pre}})$ is an estimator of the uncertainty of the predictions.
This definition enables the score function to capture the variability of the uncertainty across the input space, and to provide a more informative uncertainty estimation through the set:
\begin{equation}
    \mathcal{S}_{\dcal} = \qty{S(\vb{x}, y),~ \qty(\vb{x}, y) \in \dcal}.
\end{equation}
This score function is used to compute the empirical quantile when applied to the calibration set:
\begin{equation}
    \label{eq:cp-quantile-score}
    \qdas{\dcal}
    =
    \hat{\quant}_{(1-\alpha)\frac{m+1}{m}}\qty(\mathcal{S}_{\dcal})
    =
    \rank_{(1-\alpha)\frac{m+1}{m}}\qty( \mathcal{S}_{\dcal} ).
\end{equation}
At test time, the quantile values in the \pint are multiplied by the uncertainty estimates, computed on the unseen sample $\vb{x}$:
\begin{equation}
    \label{eq:cp-prediction-score}
    \cdas{\dtr}(\vb{x})
    =
    \left[
        \mu(\vb{x}; \hat{\theta}_{\text{pre}}) - \sigma(\vb{x}; \hat{\theta}_{\text{pre}})\; \qdas{\dcal},~
        \mu(\vb{x}; \hat{\theta}_{\text{pre}}) + \sigma(\vb{x}; \hat{\theta}_{\text{pre}})\; \qdas{\dcal}
        \right].
\end{equation}
The estimator $\sigma(\vb{x}; \hat{\theta}_{\text{pre}})$ can be obtained using several approaches.
In this work, a \mlp with batch normalisation is trained on the absolute residuals of the pretraining set $\dptr$, providing an unbiased estimate for the uncertainty of the \chf prediction.

\subsection{Heteroscedastic Regression}\label{subsec:method-hr}

\hr provides a unified framework to optimise the predictions and estimate the corresponding aleatoric uncertainties jointly.
Given a heteroscedastic model, the goal is to learn both the mean $\mu$ and the standard deviation $\sigma$ (aleatoric uncertainty) of the observed values of \chf, since they both depend on the input features $\vb{x}$.
To predict both $\mu$ and $\sigma$ simultaneously, it is necessary to modify the regression models described in Section~\ref{subsec:method-resnet}, by using an output layer that provides two distinct quantities.
The loss function used in this approach is a generalised version (generalised \hr, or \ghr) of the negative log-likelihood (\nll) Gaussian loss:
\begin{equation}
    \label{eq:hr-generalized-loss}
    {L}_{\text{\ghr}} \left(\theta; X, \vb{y}, \gamma \right)
    =
    \frac{1}{N}
    \sum_{i=1}^{N}
    \left(
    \frac{\gamma}{2} \ln \left( \sigma^2(\vb{x}_i; \theta) \right) + \left(1 - \gamma\right) \frac{{\left( y_i - \mu(\vb{x}_i; \theta) \right)}^2}{2 \sigma^2(\vb{x}_i; \theta)}
    \right),
\end{equation}
where $\gamma \in \left(0, 1\right)$ is a weight hyperparameter.
Though the uncertainty bounds should be computed using the prediction uncertainty, here we assume that the model uncertainty is negligible (the epistemic or systematic component), an assumption later validated by the Bayesian analysis in Section~\ref{subsec:results-bhr}.
Therefore we calculate the width of the bounds by only using the predicted standard deviation $\sigma( \vb{x}; \hat{\theta}_{\text{tr}})$, computed with the model trained on $\dtr$, as:
\begin{equation}
    \label{eq:hr-uncertainty-bounds}
    \cda{\dtr}\qty(\vb{x})
    =
    \left[
        \mu \left( \vb{x}; \hat{\theta}_{\text{tr}} \right) - \hat{\delta}_{\dtr}^{(\alpha)} \left( \vb{x} \right),~
        \mu \left( \vb{x}; \hat{\theta}_{\text{tr}} \right) + \hat{\delta}_{\dtr}^{(\alpha)} \left( \vb{x} \right)
        \right],
\end{equation}
where we define:
\begin{equation}
    \label{eq:hr-uncertainty-bound}
    \hat{\delta}_{\dtr}^{(\alpha)} \left( \vb{x} \right)
    =
    \sigma \left( \vb{x}; \hat{\theta}_{\text{tr}} \right) \cdot \Phi^{-1} \left(1 - \frac{\alpha}{2} \right),
\end{equation}
with $\Phi^{-1}$ being the inverse cumulative distribution function, here supposed to be Gaussian.
Therefore, the weight factor $\gamma$ enables regulation of the robustness of the model by adjusting the task of minimisation of the standard deviation with respect to the minimisation of the prediction error.
In particular, using $\gamma \in (0, 0.5)$ favours more robustness, leading to larger uncertainty, whereas choosing $\gamma \in (0.5, 1)$ gives more importance to the prediction objective, allowing for narrower bounds.
By means of this generalisation, we allow the model to deal with a possible non-Gaussian behaviour of the CHF, and to capture a representative uncertainty value at training time.
Note that assuming $\gamma = 0.5$ is equivalent to using the unweighted \nll loss.

\begin{figure}[t]
    \centering
    \begin{tikzpicture}
        \fill[dartmouthgreen!5] (0, 0) rectangle (2.2, 3.1);
        \fill[dartmouthgreen!80] (0, 0) rectangle (0.2, 2.7);
        \fill[dartmouthgreen!80] (0.2, 2.7) rectangle (2.2, 3.1);

        \foreach \i in {0.6, 1.0, 1.4, 1.8} {
                \draw[thin, black!15] (\i, 0) -- (\i, 3.1);
            }
        \foreach \i in {2.4, 2.1, 1.8, 1.5, 1.2, 0.9, 0.6, 0.3} {
                \draw[thin, black!15] (0, \i) -- (2.2, \i);
            }
        \draw[thick] (0, 2.7) -- (2.2, 2.7);
        \draw[thick] (0.2, 0) -- (0.2, 3.1);

        \node[above] at (1.2, 3.1) {\textsc{Input Data}};
        \node[anchor=center] (X) at (1.2, 1.35) {\huge $\vb{x}$};

        \draw[cobalt!80, -{Latex}, thick] (2.2, 1.35) -- (3.2, 1.35) node (arrow1) {};
        \node[draw=cobalt!80, fill=cobalt!5, text=cobalt, right=of arrow1, xshift=-3em, rectangle, inner sep=5pt] (backbone) {\large \textsc{Backbone}};
        \draw[cobalt!80, -{Latex}, thick] (backbone.east) -- ++(1.0, 0) node (arrow2) {};
        \node[draw=cobalt!80, fill=cobalt!5, text=cobalt, right=of arrow2, xshift=-3em, rectangle, inner sep=5pt] (joint) {\large \textsc{Joint Layer}};

        \draw[cobalt!80, -{Latex}, thick] (joint.east) -- ++(0.5, 0) node (centernode) {} -- ++(0, 1.0) -- ++(0.5, 0) node[right, cobalt] {\Large $\mu(\vb{x}; \theta)$};
        \draw[cobalt!80, -{Latex}, thick] (centernode.center) -- ++(0, -1.0) -- ++(0.5, 0) node[right, cobalt] {\Large $\sigma(\vb{x}; \theta)$};
    \end{tikzpicture}
    \caption{Architecture of the double-head network for \hr for predicting the \chf value $\mu(\vb{x}; \theta)$ and its uncertainty $\sigma(\vb{x}; \theta)$.}\label{fig:hr-double-head-network}
\end{figure}
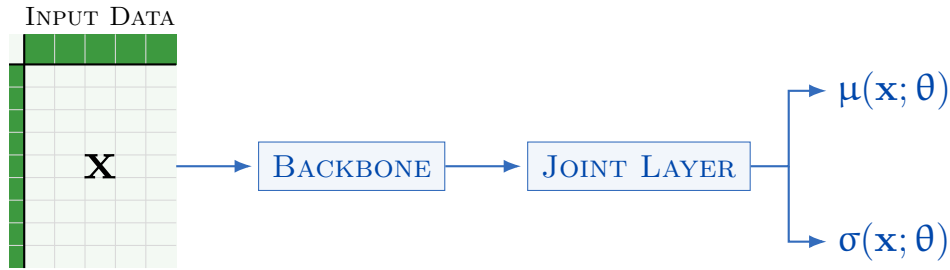

Since the model must estimate a double output, a possible solution is to use multi-task learning (\mtl) to learn $\mu$ and $\sigma$ by means of a shared representation.
This approach is implemented by designing a neural network made of a unique backbone $f_{\omega}$ and two different prediction heads $g_{\alpha_{1}}^1$ and $g_{\alpha_{2}}^2$, with a joint layer to project the output of the backbone to the space of the prediction heads, as shown in Figure~\ref{fig:hr-double-head-network}.
In this case, the model parameters are
\begin{equation}
    \theta = \omega \cup \alpha_1 \cup \alpha_2,
\end{equation}
where $\omega$ are the parameters of the backbone, and $\alpha_1$ and $\alpha_2$ are the parameters of the prediction heads for $\mu$ and $\sigma$, respectively.
However, as the quantities $\mu(\vb{x}; \theta)$ and $\sigma(\vb{x}; \theta)$ are employed in the same loss function~\eqref{eq:hr-generalized-loss}, using two different prediction heads could introduce a spurious independence, which may hamper the learning process.
This could cause the multi-task network to obtain worse results on \chf prediction than the single-task model.
In this article, we, nonetheless, explore this approach, as it allows the model to learn a shared representation for both tasks, which may be beneficial for the learning process and the generalisation capability of the model.

Since the model architecture has not been substantially modified from the $\mu$-prediction-only architecture of Section~\ref{subsec:method-resnet}, as an alternative, it is possible to employ a transfer learning (\tl) approach.
In this case, we initialise the model with the pretrained weights of the corresponding regression model and simply replace the output layer to output two scalar quantities.
\tl is often used to improve performance on a task in a domain, leveraging the knowledge acquired from an upstream domain and task.
This technique is widely considered good strategy to achieve good performance on a task when few data points are available~\cite{transfer-learning-survey}.
In this case, we are working in an inter-task setting, thus using the same dataset to train both models.
In what follows, we load the pretrained weights of all layers of the regression model, except for the output layer, which is the only architectural difference between tasks.
As the deepest layers of the \nn have already been trained in the upstream task, they are expected to have learned useful representations for the downstream task as well: we thus freeze their weights during training, leaving only the weights of the last layers trainable.
In this work, we explore both \mtl and \tl strategies for \hr, highlighting their respective advantages and disadvantages in the following sections.

\subsection{Quality-Driven Prediction and Uncertainty Estimation}\label{subsec:method-qd}

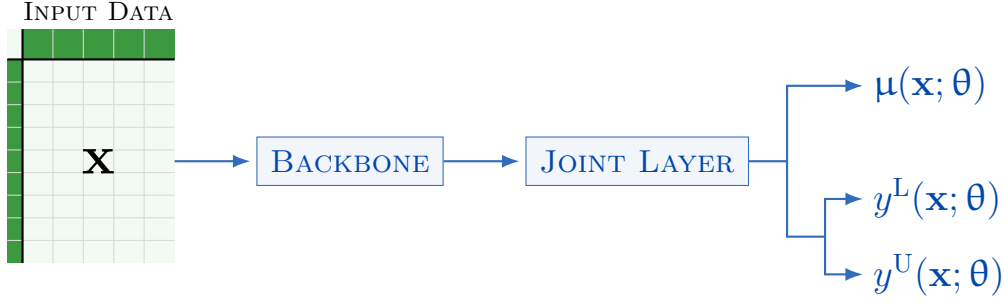
\begin{figure}[t]
    \centering
    \begin{tikzpicture}
        \fill[dartmouthgreen!5] (0, 0) rectangle (2.2, 3.1);
        \fill[dartmouthgreen!80] (0, 0) rectangle (0.2, 2.7);
        \fill[dartmouthgreen!80] (0.2, 2.7) rectangle (2.2, 3.1);

        \foreach \i in {0.6, 1.0, 1.4, 1.8} {
                \draw[thin, black!15] (\i, 0) -- (\i, 3.1);
            }
        \foreach \i in {2.4, 2.1, 1.8, 1.5, 1.2, 0.9, 0.6, 0.3} {
                \draw[thin, black!15] (0, \i) -- (2.2, \i);
            }
        \draw[thick] (0, 2.7) -- (2.2, 2.7);
        \draw[thick] (0.2, 0) -- (0.2, 3.1);

        \node[above] at (1.2, 3.1) {\textsc{Input Data}};
        \node[anchor=center] (X) at (1.2, 1.35) {\huge $\vb{x}$};

        \draw[cobalt!80, -{Latex}, thick] (2.2, 1.35) -- (3.2, 1.35) node (arrow1) {};
        \node[draw=cobalt!80, fill=cobalt!5, text=cobalt, right=of arrow1, xshift=-3em, rectangle, inner sep=5pt] (backbone) {\large \textsc{Backbone}};
        \draw[cobalt!80, -{Latex}, thick] (backbone.east) -- ++(1.0, 0) node (arrow2) {};
        \node[draw=cobalt!80, fill=cobalt!5, text=cobalt, right=of arrow2, xshift=-3em, rectangle, inner sep=5pt] (joint) {\large \textsc{Joint Layer}};

        \draw[cobalt!80, -{Latex}, thick] (joint.east) -- ++(0.5, 0) node (centernode) {} -- ++(0, 1.0) -- ++(1.0, 0) node[right, cobalt] {\Large $\mu(\vb{x}; \theta)$};
        \draw[cobalt!80, -{Latex}, thick] (centernode.center) -- ++(0, -1.0) -- ++(0.5, 0) node (mtlnode) {} -- ++(0, 0.5) -- ++(0.5, 0) node[right, cobalt] {\Large $y^{\text{L}}(\vb{x}; \theta)$};
        \draw[cobalt!80, -{Latex}, thick] (mtlnode.center) -- ++(0, -0.5) -- ++(0.5, 0) node[right, cobalt] {\Large $y^{\text{U}}(\vb{x}; \theta)$};
    \end{tikzpicture}
    \caption{Architecture of the multitask network for \qd to predict the \chf value $\mu(\vb{x}; \theta)$ and its uncertainty bounds $y^{\text{L}}(\vb{x}; \theta)$ and $y^{\text{U}}(\vb{x}; \theta)$.}\label{fig:qd-three-head-network}
\end{figure}

This method enables to optimise the \chf prediction and the value of the upper and lower uncertainty bounds, separately: one output of the network predicts the value of the \chf, while two additional outputs predict the lower and upper uncertainty bounds, respectively (see Figure~\ref{fig:qd-three-head-network}).
It allows the model to find asymmetric bounds, unlike \cpred and \hr, in an end-to-end pipeline.
The \nn used to implement this approach is similar to that employed for \hr, with output heads.
Therefore, \mtl and \tl can be analogously utilised in this context.
This method also grants a coverage of at least $1 - \alpha$ during the training process via the addition of a constraint in the loss function:
\begin{equation}
    \label{eq:qd-loss-complete}
    L(\theta; X, \vb{y}, \alpha, \gamma, \lambda) = \gamma\, L_{\text{\rmspe}}(\theta; X, \vb{y}) + \left( 1 - \gamma \right)\, L_{\text{\qd}}(\theta; X, \vb{y}, \alpha, \lambda),
\end{equation}
where $\gamma$ is a weight factor used to mitigate the different scales of the two loss terms, and $L_{\text{\rmspe}}$ is the \rmspe loss, described in~\eqref{eq:rmspe-loss}.
$L_{\text{\qd}}$ is the \emph{quality-driven} loss, defined by~\cite{qd-loss} and expressed as:
\begin{equation}
    \label{eq:qd-loss}
    L_{\text{\qd}}(\theta; X, \vb{y}, \alpha, \lambda) = L_{\text{\mpiw}}(\theta; X, \vb{y}) + \lambda\, \frac{n}{\alpha \left(1 - \alpha\right)}\, {\max \left(0,  \left(1 - \alpha\right) - \operatorname{\picp}(\theta; X, \vb{y}) \right)}^2,
\end{equation}
where $\lambda$ is a hyperparameter controlling the strength of the penalising term, and $L_{\text{\mpiw}}$ aims at minimising the average width of the uncertainty bounds of the enveloped data points:
\begin{equation}
    \label{eq:qd-loss-mpiw-loss}
    L_{\text{\mpiw}}(\theta; X, \vb{y})
    =
    \frac{1}{M} \sum_{i=1}^{n} \left(%
    y^{\text{U}}(\vb{x}_i; \theta) - y^{\text{L}}(\vb{x}_i; \theta)
    \right)
    \cdot
    \mathbb{I}(\vb{x}_i, y_i; \theta).
\end{equation}
Here, $y^{\text{U}}$ and $y^{\text{L}}$ are the upper and lower bounds, respectively, which the network predicts, and $\mathbb{I}$ is the indicator function:
\begin{equation}
    \mathbb{I}(\vb{x}_i, y_i; \theta) =
    \begin{cases}
        1 & \qif{y^{\text{L}}(\vb{x}_i; \theta) \le y_i \le y^{\text{U}}(\vb{x}_i; \theta)} \\
        0 & \quad \text{otherwise}
    \end{cases}.
\end{equation}
Using this notation, we have:
\begin{equation}
    M = \sum_{i=1}^{n} \mathbb{I}(\vb{x}_i, y_i; \theta),
\end{equation}
that is the number of enveloped samples in the current mini-batch of $n$ samples.
The second term of~\eqref{eq:qd-loss} accounts for the coverage constraint, by imposing a penalty when the prediction interval coverage probability (\picp), defined by:
\begin{equation}
    \label{eq:loss-picp}
    \operatorname{\picp}(\theta; X, \vb{y}) = \frac{1}{n} \sum_{i=1}^{n} \mathbb{I}(\vb{x}_i, y_i; \theta),
\end{equation}
is smaller than the desired coverage $1 - \alpha$.
Finally, since the indicator function is a rectangular function, thus flat almost everywhere, a smooth approximation is used~\cite{qd-loss}:
\begin{equation}
    \label{eq:qd-loss-k-soft}
    \mathbb{I}^{\text{(soft)}}(\vb{x}, y; \theta, s) = \Sigma \left( s \cdot \left( y - y^{\text{L}}(\vb{x}; \theta) \right) \right) \odot \Sigma \left( s \cdot \left( y^{\text{U}}(\vb{x}; \theta) - y \right) \right).
\end{equation}
Here, $\Sigma$ represents the sigmoid function~\cite{sigmoid}, $s > 0$ its smoothing factor, and $\odot$ indicates the element-wise multiplication.
This product of sigmoids best approximates the rectangular function as $s$ becomes larger.
One possible drawback of the approach is that, at the same time, $\mathbb{I}^{\text{(soft)}}$ also becomes flatter, leading to a possible vanishing gradient phenomenon, where the neural network struggles to learn.
The choice of a residual architecture (Section~\ref{subsec:method-resnet}) is particularly critical here to mitigate the vanishing gradient issues inherent to the smooth approximations required for the quality-driven loss.
Techniques to mitigate these issues have been discussed in Section~\ref{subsec:method-resnet}.

\subsection{Bayesian Heteroscedastic Regression}\label{subsec:method-bhr}

In this work, \bhr consists of the implementation of \hr by using a \bnn.
\bnn architectures consider the set of weights $\theta$ of a \nn as random variables, following a probability distribution, rather than a realisation, that is a specific choice, of the parameters.
This enables the separate capture of both the model (i.e.\ the systematic or epistemic) and the aleatoric component of the prediction uncertainty.
We consider an arbitrary posterior distribution of the parameters:
\begin{equation}
    \label{eq:bnn-params-posterior}
    P \left( \phi | \dtr \right) \propto P \left( \dtr | \phi \right) P \left( \phi \right),
\end{equation}
where $P \left( \phi \right) \sim \mathcal{N} \left(0, \sigma_{\phi}^2\right)$ represents the known prior distribution and $P \left( \dtr | \phi \right)$ indicates the likelihood of the training data $\dtr$, given the model parameters $\phi$.
The posterior predictive distribution, used to perform inference on an unseen sample $\left(\vb{x}, y\right)$ is:
\begin{equation}
    \label{eq:bnn-posterior-predictive}
    P \left( y | \vb{x}, \dtr \right)
    =
    \int \dd{\phi} P \left( y | \vb{x}, \phi \right) P \left( \phi | \dtr \right)
    =
    \mathbb{E}_{\phi \sim P \left( \phi | \dtr \right)} \left[ P \left( y | \vb{x}, \phi \right) \right],
\end{equation}
where $P \left( y | \vb{x}, \phi \right)$ represents the probability distribution of the generic new observation $y$ given the input features $\vb{x}$ and the model parameters $\phi$~\cite{bnn-def}.
In a \bnn, the prediction is obtained by averaging multiple realisations of the posterior distribution.
This model is optimised using variational inference-based methods, by minimising the Kullback-Leibler (\kl) divergence $D_{\text{\kl}}$ between an approximate distribution of the parameters $Q \left(\phi | \theta \right)$, depending on its own set of variational parameters $\theta$, and the true posterior distribution $P \left( \phi | \mathcal{D}_{tr} \right)$~\cite{bnn-bayes-by-backprop}:
\begin{equation}
    \begin{split}
        \hat{\theta}
        & =
        \arg\min\limits_{\theta} D_{\text{\kl}} \left( Q \left(\phi | \theta \right) \Vert P \left( \phi | \dtr \right) \right) \\
        & =
        \arg\min\limits_{\theta}
        \int \dd{\phi} Q \left(\phi | \theta \right) \ln \frac{Q \left(\phi | \theta \right)}{P \left( \dtr | \phi \right) P \left( \phi \right)} \\
        & =
        \arg\min\limits_{\theta}
        \int \dd{\phi} Q \left(\phi | \theta \right) \ln \frac{Q \left(\phi | \theta \right)}{P \left( \phi \right)}
        -
        \int \dd{\phi} Q \left(\phi | \theta \right) \ln P \left( \dtr |\phi \right) \\
        & =
        D_{\text{\kl}} \left( Q \left(\phi | \theta \right) \Vert P \left( \phi \right) \right)
        -
        \mathbb{E}_{\phi \sim Q \left( \phi | \theta \right)} \left[ \ln P \left( \dtr | \phi \right) \right].
    \end{split}
\end{equation}
The \emph{reparametrisation trick} is used to train the \bnn with backpropagation, while ensuring a certain degree of flexibility of the variational distribution $Q \left(\phi | \theta \right)$~\cite{reparametrization-trick}.
The $\beta$-Negative \emph{evidence lower bound} (\elbo) loss is used to optimise these architectures:
\begin{equation}
    \label{eq:bnn-beta-neg-elbo-loss}
    L_{\beta\text{-\bnn}} \left( \theta; \dtr \right)
    =
    - \mathbb{E}_{Q \left( \phi | \theta \right)} \left[ \ln P \left( \dtr | \phi \right) \right]
    +
    \beta_{\text{\kl}}\, D_{\text{\kl}} \left( Q \left( \phi | \theta \right) \Vert P \left( \phi\right) \right),
\end{equation}
where the first term coincides with the \nll loss~\eqref{eq:hr-generalized-loss} and $\beta_{\text{\kl}}$ is a weight factor used to balance the prediction optimisation and the regularisation provided by the second term.

\uq is performed as described by~\cite{qd-loss}, by feeding the model with $N$ instances of the same sample.
The predicted mean $\mu$ is computed as the average of the posterior predictive distribution, which can be approximated by sampling multiple times from the posterior distribution of the parameters:
\begin{equation}
    \label{eq:bnn-predicted-value}
    \mu(\vb{x}; \hat{\theta})
    =
    \mathbb{E}_{y \sim P\qty(y | \vb{x}, \dtr)}\qty[%
        \mathbb{E}_{\theta \sim P \left( \theta | \dtr \right)} \left[ P \left( y | \vb{x}, \theta \right) \right]
    ].
\end{equation}

The estimator of the aleatoric uncertainty $\sigma_{\text{data}}^2$ is calculated as the average of the variance values obtained at each iteration.
The model uncertainty $\sigma^2_{\text{model}}$ can be reasonably approximated as the sample variance of the predicted mean values, for a number of samples sufficiently large.
The total prediction uncertainty is then computed as:
\begin{equation}
    \begin{split}
        \sigma_{\text{pred}}^2(\vb{x}; \hat{\theta})
        & =
        \operatorname{Var}_{y }\qty(%
        \mathbb{E}_{\theta } \qty[P\qty(y \mid \vb{x}, \theta)]
        )
        +
        \mathbb{E}_{y}\qty[%
            \operatorname{Var}_{\theta}\qty(P\qty(y \mid \vb{x}, \theta))
        ] \\
        & \approx
        \sigma_{\text{model}}^2(\vb{x}; \hat{\theta}) + \sigma_{\text{data}}^2(\vb{x}; \hat{\theta})
    \end{split}
\end{equation}
where the approximation comes from the model uncertainty.
Uncertainty bounds are calculated as in \hr, using~\eqref{eq:hr-uncertainty-bounds} and~\eqref{eq:hr-uncertainty-bound}.

\section{Evaluation metrics}\label{sec:eval-metrics}

In this section we introduce the metrics used to evaluate the goodness-of-fit of \ml models and their associated uncertainty.
In particular, we consider separately the evaluation of the performance of the \ml models via the evaluation of the distance from the ground truth value of the \chf.
We then introduce metrics to evaluate coverage and quality of the \uq.

While training aims to determine the optimal parameters $\hat{\theta}$ of the \ml architecture through minimisation of the loss functions defined in Section~\ref{sec:method}:
\begin{equation}
    \hat{\theta}
    =
    \arg\min\limits_{\theta} L(\theta; \dtr, \Omega),
\end{equation}
where $\Omega$ is the set of hyperparameters, evaluation is carried out on predictions generated with $\hat{\theta}$.
This justifies the slight change in notation of certain metrics, which may look similar to some loss functions, though they explicitly depend on fixed optimal parameters $\hat{\theta}$, rather than the generic $\theta$ used during training.

\subsection{Machine Learning Metrics}\label{subsec:eval-metrics-ml}

We take inspiration from the metrics used for \ml models defined in the context of the benchmark~\cite{benchmark}.
Given the $i$-th predicted and observed (ground truth) \chf values of $N$ samples, these metrics are the \emph{mean absolute percentage error} (\mape), the \rmspe (already defined as a loss function in previous sections), and the $Q^2$ error (detailed later) expressed as:
\begin{equation}
    \label{eq:metric-mape}
    \operatorname{\mape}(X, \vb{y}; \hat{\theta}) = 100 \cdot \frac{1}{N} \sum_{i=1}^{N} \left| \frac{y_{i} - \mu(\vb{x}_i; \hat{\theta})}{y_{i}} \right|,
\end{equation}
\begin{equation}
    \label{eq:metric-rmspe}
    \operatorname{\rmspe}(X, \vb{y}; \hat{\theta}) = 100 \cdot \sqrt{ \frac{1}{N} \sum_{i=1}^{N} {\left(\frac{y_{i} - \mu(\vb{x}_i; \hat{\theta})}{y_{i}}\right)}^{2} },
\end{equation}
\begin{equation}
    \label{eq:metric-q2}
    Q^2(X, \vb{y}; \hat{\theta}) = \frac{\sum_{i} {(\mu(\vb{x}_i; \hat{\theta}) - y_{i})}^2}{\sum_{i} {(\mu(\vb{x}_i; \hat{\theta}) - \bar{y})}^2},
\end{equation}
where $\bar{y}$ is the experimental mean of the \chf values in the dataset.
The $Q^2$ \emph{error} measures the amount of variability of the experimental data explained by the \ml model.
A value greater than 1 indicates that the model is worse than a dummy model that always predicts the experimental mean $\bar{y}$.
The \mlp model used in adaptive \cpred, described in Section~\ref{subsec:method-cp}, is instead validated using the \rmse:
\begin{equation}
    \label{eq:metric-rmse}
    \operatorname{\rmse}(X, \vb{y}; \hat{\theta}) = \sqrt{\frac{1}{N} \sum_{i=1}^{N}{(y_i - \mu(\vb{x}_i; \hat{\theta}))}^2}.
\end{equation}
This is due to the fact that the \mlp in the upstream task was found to perform best when trained to minimise the \mse loss, rather than the \rmspe.

The calibration of the \ml models is then assessed using the \emph{Area Under Calibration Error} curve (\auce).
It compares the experimental and predicted quantiles to quantify the overall distance between predictions and ground truths~\cite{calibration-curves}.
In particular, calling $\Phi^{-1}$ and $\hat{\Phi}^{-1}$ the experimental and predicted quantiles (i.e.\ the inverse functions of the respective cumulative distribution functions), the \auce is computed as:
\begin{equation}
    \label{eq:calibration-auce}
    \operatorname{\auce}(X, \vb{y}; \hat{\theta})
    =
    \int_{0}^{1} \dd{p}\, \left| \hat{\Phi}^{-1}(p, X; \hat{\theta}) - \Phi^{-1}(p, \vb{y}) \right|;
\end{equation}
An \auce of 0 represents the ideal case of perfect calibration, when all the predicted quantiles coincide with the corresponding experimental ones.

\subsection{Uncertainty Quantification Metrics}\label{subsec:eval-metrics-uq}

\uq is generally non-trivial when applied to \ml models, since different metrics, learning objectives, and architectures can be used.
A useful metric, in what follows, is the \picp~\cite{qd-loss}, which measures the fraction of experimental data enveloped by the uncertainty bounds:
\begin{equation}
    \label{eq:metric-picp}
    \operatorname{\picp}(X, \vb{y}; \hat{\theta}) = \frac{1}{N} \sum_{i=1}^{N} \mathbb{I} \left[ y^{\text{L}}(\vb{x}_i; \hat{\theta}) \le y_{i} \le y^{\text{U}}(\vb{x}_i; \hat{\theta}) \right],
\end{equation}
where $y^{\text{L}}(\vb{x}; \hat{\theta})$, $y^{\text{U}}(\vb{x}; \hat{\theta})$ are the lower and upper predicted uncertainty bounds, respectively, $y$ is the ground truth value of the \chf, $\mathbb{I}[e]$ is the usual indicator function introduced for~\eqref{eq:loss-picp} ($1$ where the statement $e$ is true, $0$ otherwise), and $N$ is the number of samples in the dataset on which this metric is computed.
Though a similar definition of the function can also be used as loss as in~\eqref{eq:loss-picp}, the \picp is here computed on the predictions obtained with the optimal parameters $\hat{\theta}$, rather than during training, to evaluate the coverage of the \uq.
In this work, \picp will be referred to as \emph{coverage} for simplicity.

Other relevant metrics are the \emph{relative informativeness} and the \emph{uncertainty calibration}, both ranging in $\left[0,1\right]$~\cite{sapium, DESTERCKE20082484}.
The former measures the relative narrowness of the uncertainty bounds for a prediction $\mu(\vb{x}; \hat{\theta})$, and it is expressed as:
\begin{equation}
    \label{eq:metric-informativeness-sample}
    \operatorname{\inform}_{i}(\vb{x}_i; \hat{\theta}) =
    \begin{cases}
        1 - \dfrac{y^{\text{U}}(\vb{x}_i; \hat{\theta}) - y^{\text{L}}(\vb{x}_i; \hat{\theta})}{2\, \mu(\vb{x}_i; \hat{\theta})} \quad & \text{if} \quad 0 \le y^{\text{U}}(\vb{x}_i; \hat{\theta}) - y^{\text{L}}(\vb{x}_i; \hat{\theta}) \le 2\, \mu(\vb{x}_i; \hat{\theta}) \\
        0                                                                                                                              & \text{otherwise}
    \end{cases},
\end{equation}
for $i = 1, 2, \dots, N$.
Notice that the metric is naturally bounded between $0$ and $1$ by imposing that relative uncertainties of more than 100\% correspond to an informativeness of zero.

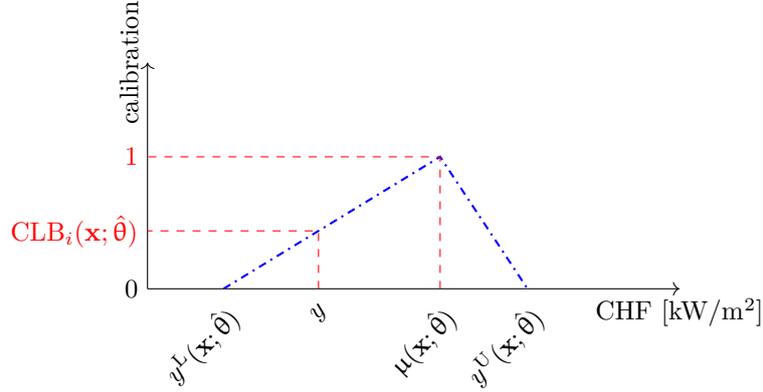
\begin{figure}[t]
    \centering
    \begin{tikzpicture}
        \draw[->] (0, 0) node[left] {$0$} -- (7, 0) node[below] {\chf $[\si{\kilo\watt\per\meter\squared}]$};
        \draw[->] (0, 0) -- (0, 3) node[above, rotate=90] {calibration};

        \draw[thick, blue, dashdotted] (1, 0) -- (3.85, 1.75) -- (5, 0);

        \node[rotate=45, anchor=north east] at (1.0, 0.0) {$y^{\text{L}}(\vb{x}; \hat{\theta})$};
        \node[rotate=45, anchor=north east] at (5.0, 0.0) {$y^{\text{U}}(\vb{x}; \hat{\theta})$};
        \node[rotate=45, anchor=north east] at (3.85, 0.0) {$\mu(\vb{x}; \hat{\theta})$};
        \node[rotate=45, anchor=north east] at (2.25, 0.0) {$y$};

        \draw[dashed, red] (2.25, 0) -- (2.25, 0.768) -- (0, 0.768) node[left] {$\operatorname{\clb}_i(\vb{x}; \hat{\theta})$};
        \draw[dashed, red] (3.85, 0) -- (3.85, 1.75) -- (0, 1.75) node[left] {1};
    \end{tikzpicture}
    \caption{Triangular distribution used for the uncertainty calibration.}\label{fig:triang-distrib-cal}
\end{figure}

Uncertainty calibration measures the compatibility of the experimental value with the prediction, given the uncertainty estimate~\cite{sapium}:
\begin{equation}
    \label{eq:metric-calibration-sample}
    \operatorname{\clb}_{i}(\vb{x}_i, y_i; \hat{\theta}) =
    \begin{dcases}
        \frac{y_{i} - y^{\text{L}}(\vb{x}_i; \hat{\theta})}{\mu(\vb{x}_i; \hat{\theta}) - y^{\text{L}}(\vb{x}_i; \hat{\theta})} & \quad \text{if} \quad y_{i} \in \left[y^{\text{L}}(\vb{x}_i; \hat{\theta}), \mu(\vb{x}_i; \hat{\theta})\right]                                      \\
        \frac{y_{i} - y^{\text{U}}(\vb{x}_i; \hat{\theta})}{\mu(\vb{x}_i; \hat{\theta}) - y^{\text{U}}(\vb{x}_i; \hat{\theta})} & \quad \text{if} \quad y_{i} \in \left[\mu(\vb{x}_i; \hat{\theta}), y^{\text{U}}(\vb{x}_i; \hat{\theta}) \right]                                     \\
        0                                                                                                                       & \quad \text{if} \quad y_{i} \notin \left[y^{\text{L}}(\vb{x}_i; \hat{\theta}), y^{\text{U}}(\vb{x}_i; \hat{\theta}) \right] \vphantom{\dfrac{A}{B}}
    \end{dcases},
\end{equation}
This metric generalises the \picp, which can only assume a binary value (either zero or one) for each sample, depending on the value of the ground truth with respect to the uncertainty bounds.
It relies on the triangular distribution assumption shown in Figure~\ref{fig:triang-distrib-cal}.

Relative informativeness and uncertainty calibration are strongly related.
For instance, if we consider a prediction error $\varepsilon_{i}(\vb{x}; \hat{\theta}) = y_{i} - \mu(\vb{x}_i; \hat{\theta})$ and the triangular distribution represented in Figure~\ref{fig:triang-distrib-cal}, the informativeness increases as the uncertainty bounds become narrower, and consequently the calibration decreases, and vice versa.
From a qualitative point of view, we observe the same tradeoff for precision and recall in binary classification.
Summarising the quality of \uq in the harmonic mean of these metrics (not unlike the F-score in the case of precision and recall) could be reasonably useful~\cite{fscore}.
Therefore, we propose an adapted \uq ``F-score'' version:
\begin{equation}
    \label{eq:metric-uqf-sample}
    \frac{1}{\operatorname{\uqf}_{i}(\vb{x}_i, y_i; \hat{\theta})}
    =
    \frac{1}{2}\qty(\frac{1}{\operatorname{\inform}_{i}(\vb{x}_i; \hat{\theta})} + \frac{1}{\operatorname{\clb}_{i}(\vb{x}_i, y_i; \hat{\theta})}).
\end{equation}
When referred to the entire dataset, these metrics are computed as the average of the quantities calculated on each sample:
\begin{equation}
    \label{eq:metric-informativeness}
    \operatorname{\inform}(X; \hat{\theta}) = \frac{1}{N} \sum_{i=1}^{N} \operatorname{\inform}_{i}(\vb{x}_i; \hat{\theta}),
\end{equation}
\begin{equation}
    \label{eq:metric-calibration}
    \operatorname{\clb}(X, \vb{y}; \hat{\theta}) = \frac{1}{N} \sum_{i=1}^{N} \operatorname{\clb}_{i}(\vb{x}_i, y_i; \hat{\theta}),
\end{equation}
\begin{equation}
    \label{eq:metric-uqf}
    \operatorname{\uqf}(X, \vb{y}; \hat{\theta}) = \frac{1}{N} \sum_{i=1}^{N} \operatorname{\uqf}_{i}(\vb{x}_i, y_i; \hat{\theta}),
\end{equation}
where $N$ is the number of data points.
For better clarity, the values of such metrics will be expressed as a percentage instead of a relative value.

\section{Results}\label{sec:results}

The methods described in the previous section are used for \chf prediction and \uq on the \nrc dataset.
For this purpose, 80\% of the samples are retained for training, 10\% for model validation and 10\% for the final evaluation.
\ml metrics refer to the latter, unless otherwise stated, while the \uq metrics refer to the full dataset.
The random sampling is performed by stratification on the data source to deal with possible unreliability due to instrumentation and technological changes across decades of some experiments.
The input features used are $G$, $P$, $D$, $L$, $\mathfrak{X}$ in all the approaches studied in this work.
In the case of \cpred, 300 data points are extracted from the training set and used for calibration.
For \ml operations, features are scaled using standard scaling with statistics of the training set.

Models are trained for \num{1500} epochs, when not differently specified.
We use early stopping, with a patience value set to avoid too expensive training operations when no improvement is observed.
The reference metric used for model selection is \rmspe, reported in~\eqref{eq:metric-rmspe}, except for the \mlp model used in adaptive \cpred, that is validated using the squared root of \mse (\rmse), defined in~\eqref{eq:metric-rmse}.
The chosen target coverage value is 95\%, thus we set $\alpha = 0.05$.
The work conducted by some of the current authors in~\cite{cassetta} is considered a starting point for the optimisation of the residual network for \chf prediction.

\subsection{Residual Networks}\label{subsec:results-resnet}

As a baseline experiment, we designed a \resnet as in~\cite{cassetta}, with the same number of neurons for each hidden layer.
For this first realisation, the objective is to train a model directly on \chf predictions, in order to obtain a fitted architecture, reusable for \uq and additional tasks.
Batch normalisation and \relu activation are employed as the first two stages of each residual block~\cite{resnet-mappings}.
Softplus activation is used after the output layer, to force the \chf predictions to be strictly positive.
This guarantees the consistency of the model output with the physical nature of the phenomenon.
The model is optimised by tuning the learning rate in the range $\left[ \num{1e-5}, \num{1e-3} \right]$, the weight decay coefficient in $\left[ 0, \num{1e-4} \right]$, and the smoothness factor of the softplus activation $\beta$ in $\left[1, 10\right]$.
The depth and width of the network are also considered as hyperparameters to be optimised (fixed for each hidden layer).

\begin{table}[t]
    \centering
    \caption{Hyperparameters and results of the \resnet architecture.}\label{tab:resnet-results}
    \begin{tabular}{@{}ccc@{}}
        \toprule
                                  & \textbf{Previous work}~\cite{cassetta} & \textbf{Current work} \\
        \midrule
        Depth                     & \num{30}                               & \num{8}               \\
        Width                     & \num{64}                               & \num{64}              \\
        Learning rate             & \num{1e-3}                             & \num{1e-3}            \\
        smoothness ($\beta$)      & ---                                    & \num{5}               \\
        Weight decay              & \num{1e-7}                             & \num{1e-7}            \\ \midrule
        \rmspe (\%)               & \num{11.0}                             & \textbf{10.1}         \\
        \mape (\%)                & \num{7.5}                              & \textbf{6.6}          \\
        Q\textsuperscript{2} (\%) & \num{0.02}                             & \textbf{0.01}         \\
        \bottomrule
    \end{tabular}
\end{table}

The chosen batch size is \num{512}, while the model is optimised using Adam. The initialisation strategy is Kaiming uniform~\cite{kaiming-init}.
Table~\ref{tab:resnet-results} shows the optimal hyperparameters and the results, using the metrics defined in Section~\ref{subsec:eval-metrics-ml}.
As we can observe, the primary difference between the optimal configuration of the current work and the previous one is given by the depth of the network.
The reason lies in the utilisation of the filtered dataset, which enables the achievement of similar results by using a simpler model.
The smoothness factor $\beta$ is not reported for the previous work since the softplus activation function was not used.

\begin{figure}[t]
    \centering
    \includegraphics[width=0.6\linewidth]{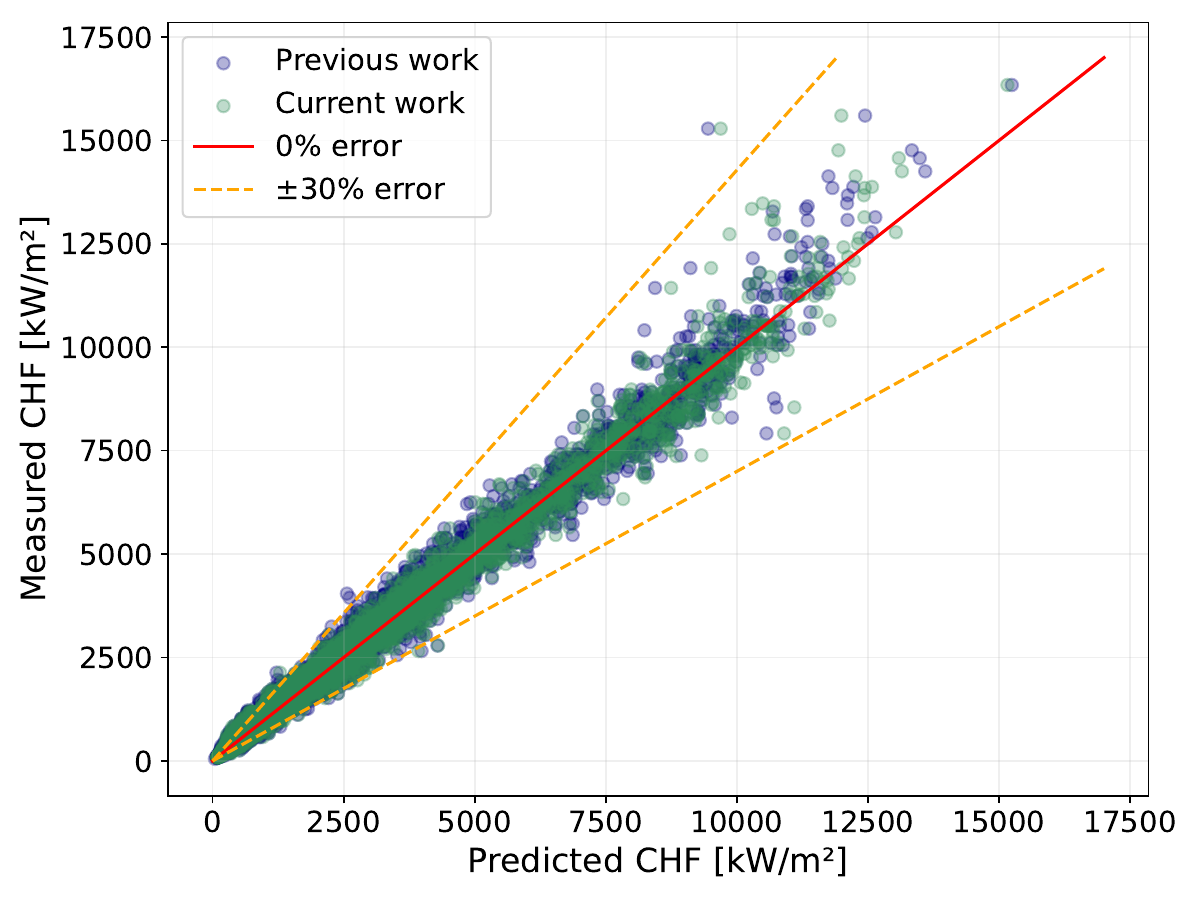}
    \caption{Measured vs predicted \chf of previous and current work.}\label{fig:resnet-scatter}
\end{figure}

Figure~\ref{fig:resnet-scatter} shows the measured and predicted values of \chf for both the previous work (in blue) and the current one (in green).
We can observe that in both cases the predictions are scattered around the 0\% error line (in red), and most of them are within the 30\% error bounds.
Qualitatively, the result is mostly unchanged from one model to the other, with neither significant underestimation nor overestimation of the predictions.
This outcome is expected since the numerical difference between the error in the two cases is less than 1\%.

The newly trained \resnet model will be used in what follows for additional downstream tasks and \uq.
Before we proceed, we test the model for heteroscedasticity with significance level $\alpha = 0.05$ to verify whether applying adaptive \cpred and \hr (and \bhr) is actually relevant.
In our case, the result of the \bp test is different from zero, with a vanishing p-value ($p \ll 0.05$).
We thus consider the \resnet trained on \chf values a heteroscedastic model.
Finally, the new \resnet model is found to be well-calibrated, by showing a low \auce of \num{0.005} on the test set.

\subsection{Conformal Prediction}\label{subsec:results-cp}

While vanilla \cpred does not require any hyperparameter tuning, the adaptive variant needs to perform model selection on the \mlp trained on the absolute residuals, called \emph{residual model} for clarity.
For the same reason, the \resnet model is here called \emph{predictive model}, to avoid confusion.
The residual model is designed with the same architectural choices as the predictive model.
The \mlp is trained for a maximum of \num{200} epochs, with early stopping (patience: \num{50} epochs), and model selection is performed using the \rmse.
A coarse optimisation has been performed automatically by using \emph{RayTune}~\cite{liaw2018tune} and the \emph{ASHA} algorithm~\cite{asha-algorithm}.
The refinement is performed manually, by tuning the network width (i.e.\ the number of neurons) in the range $[32,\, 64]$, the learning rate in $[\num{1e-3},\, \num{1e-2}]$, and the softplus smoothness factor $\beta$ in $[2,\, 10]$.
We designed a small network with 4 hidden layers, each one with 32 neurons.
Training is conducted using AdamW~\cite{adamw}, due to its better generalisation properties, with a learning rate of \num{1e-2}, and a weight decay of \num{1e-4}.
The softplus activation smoothness factor in the output layer is set to \num{10}.

\begin{table}[t]
    \centering
    \caption{Comparison between vanilla and adaptive \cpred.}\label{tab:cp-results}
    \begin{tabular}{@{}ccccc@{}}
        \toprule
        \multirow{2}{*}{\textbf{Variant}} & \multicolumn{2}{c}{\textbf{Bounds} (\%)} & \multicolumn{2}{c}{\textbf{\picp} (\%)}                                             \\
        \cmidrule(ll){2-5}
                                          & \textbf{Mean}                            & \textbf{Std}                            & \textbf{Test set} & \textbf{Full dataset} \\
        \midrule
        Vanilla                           & 31.4                                     & 30.5                                    & 93.1              & 95.1                  \\
        Adaptive                          & 22.8                                     & 16.9                                    & 96.9              & 98.6                  \\
        \bottomrule
    \end{tabular}
\end{table}

\begin{figure}[t]
    \centering
    \includegraphics[width=0.6\linewidth]{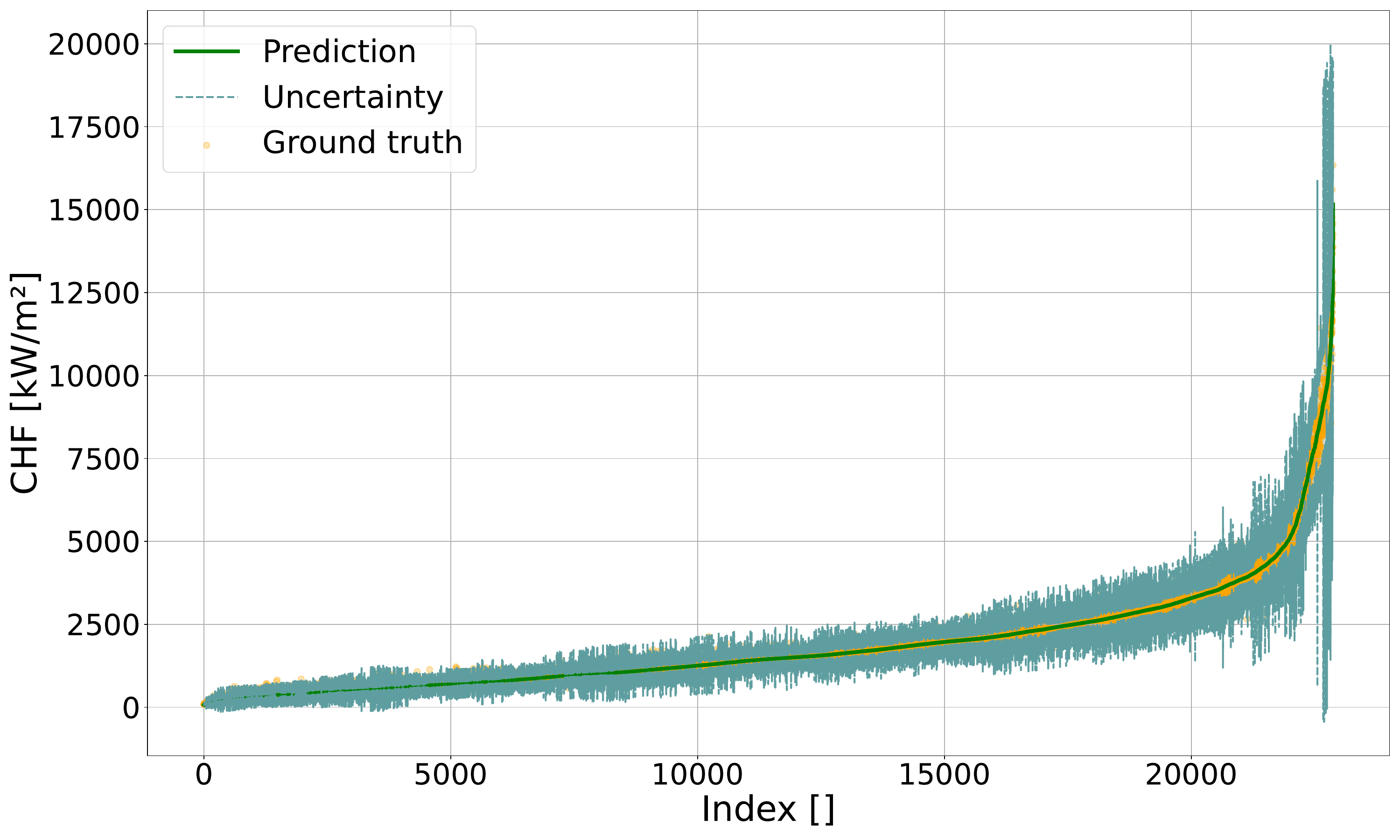}
    \caption{Uncertainty bounds of adaptive \cpred.}\label{fig:enveloping-cp-adaptive}
\end{figure}

Table~\ref{tab:cp-results} shows the result of both \cpred variants applied to the \resnet predictive model.
The adaptive method provides, on average, much narrower and less variable uncertainty bounds.
Moreover, the coverage provided by the adaptive \cpred is larger than the target value of 95\% on both the test set and the whole dataset.
The constant bounds (computed value: \SI{302}{\kilo\watt\per\meter\squared}) found by the vanilla variant are not able to achieve the desired \picp value due to their inability to adapt to the different conditions that may occur in the input features.
For the same reason, the relative width of the bounds is about 40\% larger than what found in the adaptive case, on average.
Figure~\ref{fig:enveloping-cp-adaptive} shows the model predictions on the filtered dataset and the corresponding uncertainty bounds for adaptive \cpred.
The experimental \chf values, in yellow, are mostly inside the varying bounds, which do not follow any specific trend with respect to the \chf values.
However, we notice that data points with higher \chf predictions tend to have wider uncertainty bounds than those with lower ones.

\begin{figure}[t]
    \centering
    \includegraphics[width=0.8\linewidth]{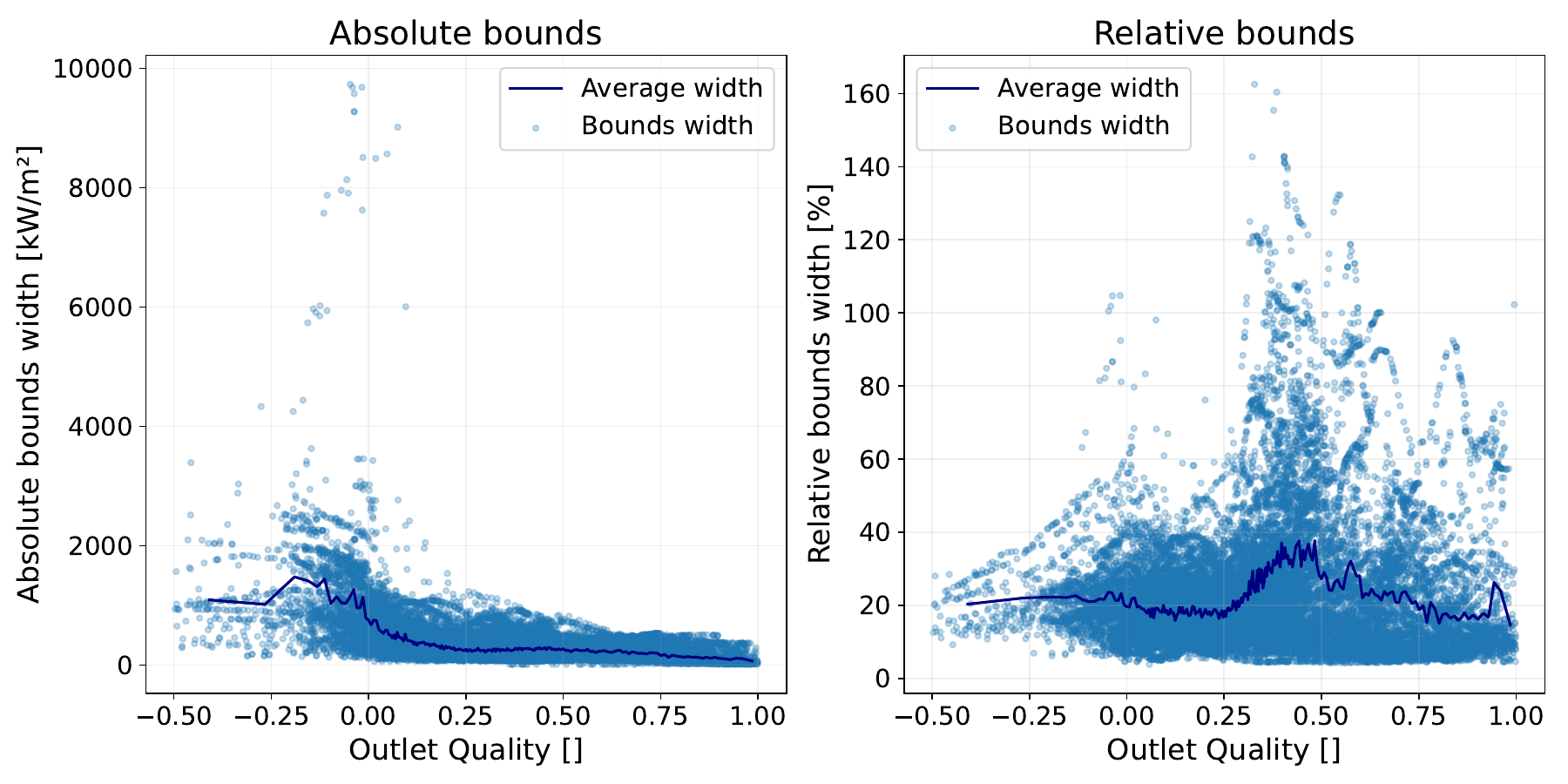}
    \caption{Uncertainty bounds vs outlet quality for adaptive \cpred.}\label{fig:cp-whole-X}
\end{figure}

\begin{table}[t]
    \centering
    \caption{\uq metrics for \cpred.}\label{tab:cp-inf-cal}
    \begin{tabular}{@{}cccccc@{}}
        \toprule
        \textbf{Variant} & \textbf{\inform} (\%) & \textbf{\clb} (\%) & \textbf{\uqf} (\%) \\
        \midrule
        Vanilla          & 70.1                  & 74.9               & 64.4               \\
        Adaptive         & 77.3                  & 74.6               & \textbf{72.5}      \\
        \bottomrule
    \end{tabular}
\end{table}

The influence of the input features on the relative width of the bounds has also been investigated, and we found that only the output quality $\mathfrak{X}$ shows a significant connection with the prediction uncertainty.
As represented on the right of Figure~\ref{fig:cp-whole-X} (relative bounds width), we can identify four intervals in the domain of $\mathfrak{X}$ with different impact on the output \chf:

\begin{enumerate}
    \item for $\mathfrak{X} < 0$, the width is mostly below 40\%, with several outliers, and it grows with $\mathfrak{X}$,
    \item where $\mathfrak{X} \in [0, 0.25]$, bounds are generally less noisy and mainly up to approximately 50\%,
    \item for the \emph{transition regime} where $\mathfrak{X} \in [0.25, 0.5]$, the amplitude spikes to more than 100\%, with high variability depending on the specific case,
    \item Points such that $\mathfrak{X} > 0.5$ show a moderate and significantly noisy width, somewhat decreasing as the outlet quality increases.\label{enum:dnb-dryout-vs-x}
\end{enumerate}

These results are consistent with the physical interpretation of the \chf.
Low $\mathfrak{X}$ values are usually associated to the \dnb regime, while \dout occurs at high values of $\mathfrak{X}$.
The prediction of these two regimes are characterised by relatively low uncertainties, while larger relative uncertainties are found in the transition regime between the two.
These observations are crucial to better understand and study the transition between the two dominant physical \chf regimes, allowing a more quantitative identification of the associated thresholds, which are usually unknown.
In this context, the transition regime starts to be quantitatively defined by the increase of the uncertanty bounds.

More specifically, we can observe some points going up to 160\% of the relative amplitude and several noisy values in all regimes.
Table~\ref{tab:cp-inf-cal} shows the results of the evaluation of \cpred using the \uq metrics.
We observe that the two approaches have nearly equivalent calibration.
In particular, adaptive \cpred has a higher \uqf score, which shows that adaptive \cpred provides more informative uncertainty bounds.

\begin{figure}[t]
    \centering
    \includegraphics[width=0.7\linewidth]{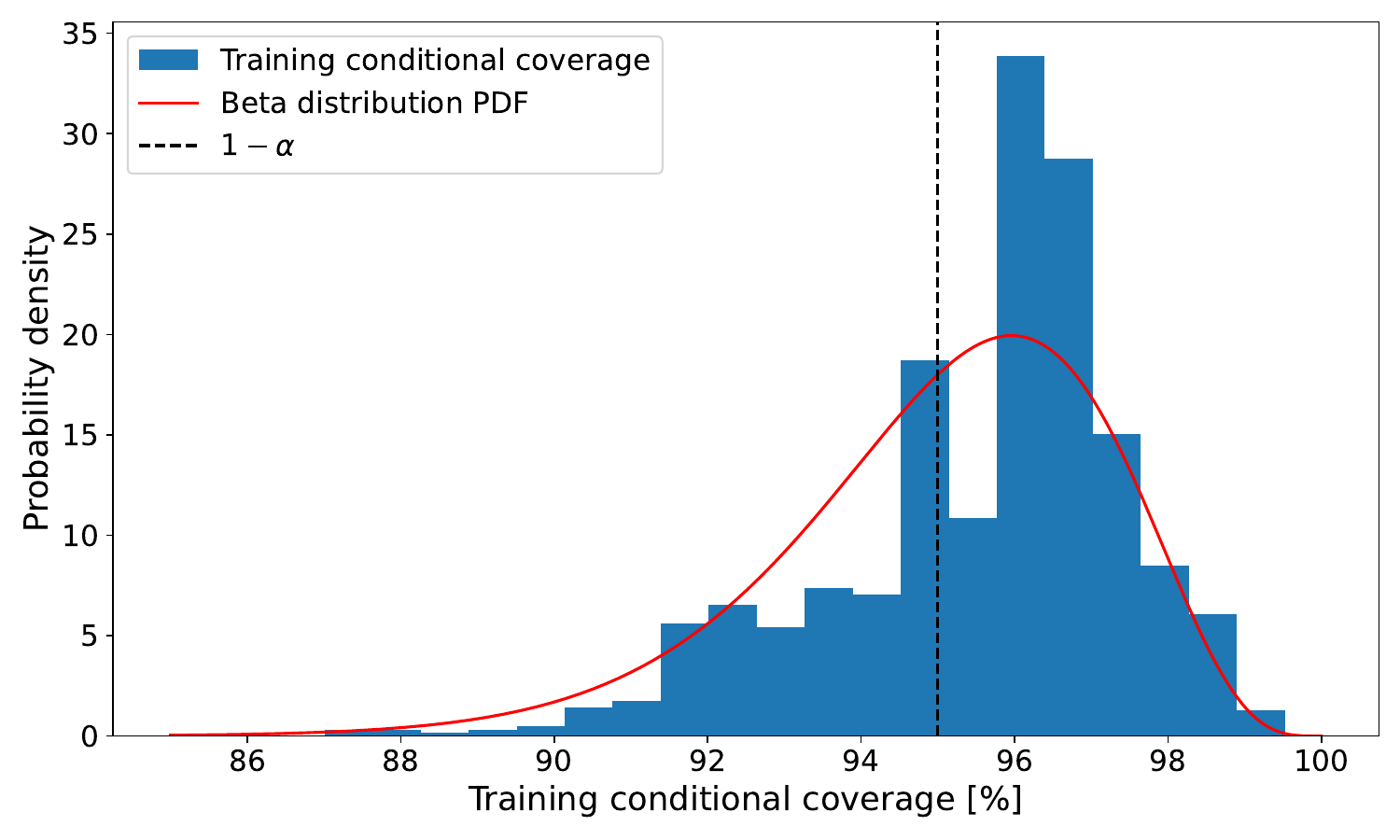}
    \caption{Beta distribution (in red) and relative frequencies of the coverage values (in blue).}\label{fig:beta-coverage}
\end{figure}

To prove the consistency of the results with the Beta distribution theorem~\eqref{eq:cp-beta-coverage}, we ran \num{10000} trials with $m = 100$ calibration samples and confidence level $1 - \alpha = 0.95$.
The data used for calibration are randomly sampled from the dataset at each iteration.
The coverage is computed on the test set, and we expect to find different values, following a Beta distribution, whose parameters depend on $m$ and $\alpha$ as described in Section~\ref{subsec:method-cp}.
Figure~\ref{fig:beta-coverage} shows the Beta distribution (in red) and discrete probability density of the coverage values (in blue), bootstrapped on the test sets.
The \picp values follow the expected Beta distribution, confirming the consistency of the results, despite being below the expected threshold $1 - \alpha$, indicated with the dashed black line.

Split \cpred can thus provide a baseline for \uq in its vanilla version, without giving any information about the influence of the input features on the uncertainty bounds.
The heteroscedasticity of the predictive model on which \cpred is applied allows us to obtain interesting results when using the adaptive variant.
In particular, we were able to detect a specific relationship between the outlet quality and the width of the bounds, compatible with a \chf expert judgement.

\subsection{Heteroscedastic Regression}\label{subsec:results-hr}

The implementation of \hr follows the same procedure employed for the baseline \resnet.
A residual network is used as \ml model, with the last layer providing a double output.
As a baseline, we designed a model similar to the one used for \chf prediction in Section~\ref{subsec:method-resnet}, where the output layer is modified to provide both the \chf prediction and the associated variance.
In this first example, the network is trained from scratch for \hr.
Variants with \mtl and \tl have also been applied, with the latter utilising only the single-task architecture provided for the \chf regression task, described in Section~\ref{subsec:results-resnet}.

The training is performed using early stopping with patience of \num{100} epochs.
Model selection is done choosing the setup with the lowest \rmspe.
Models that fail to achieve the target coverage $1-\alpha$ at evaluation time are simply discarded, even if they yield better \chf predictions.
Hyperparameter tuning mainly involves the depth of the network in the space $\left[ 8, 16 \right]$, the weight factor $\gamma$ in the range $\left[ 0.3, 0.5 \right]$, for \mtl the width of the prediction heads in $\left[ 32, 64 \right]$, and for \tl, the number of layers to freeze.
Other hyperparameters, such as the learning rate, weight decay, and the choice of the optimiser, have been chosen based on the experience detailed in the previous sections for simplicity.

\begin{table}[t]
    \centering
    \caption{Hyperparameters and results of different variants of \hr.}\label{tab:hr-results}
    \begin{tabular}{@{}cccccccc@{}}
        \toprule
        \multirow{2}{*}{\textbf{Variant}}
             & \multirow{2}{*}{\textbf{Depth}}
             & \multirow{2}{*}{$\boldsymbol{\gamma}$}
             & \multirow{2}{*}{\textbf{\rmspe} (\%)}
             & \multicolumn{2}{c}{\textbf{Bounds} (\%)}
             & \multicolumn{2}{c}{\textbf{\picp} (\%)}                                                                                              \\
        \cmidrule(ll){5-8}
             &                                          &     &               & \textbf{Mean} & \textbf{Std} & \textbf{Test set} & \textbf{Dataset} \\
        \midrule
        Base & 12                                       & 0.4 & 10.8          & 20.8          & 12.2         & 97.2              & 99.1             \\
        \mtl & 11 + 0                                   & 0.4 & 10.6          & 19.7          & 12.6         & 96.7              & 98.9             \\
        \tl  & 8                                        & 0.4 & \textbf{10.3} & 39.6          & 1.5          & 99.2              & 99.4             \\
        \bottomrule
    \end{tabular}
\end{table}

Table~\ref{tab:hr-results} shows the hyperparameters and the results of the three \hr variants.
The models are optimised using Adam (learning rate: \num{1e-3}; weight decay: \num{1e-4}).
Regarding the \mtl architecture, the depth of the network is expressed as $\operatorname{depth}_{\text{backbone}} + \operatorname{depth}_{\text{heads}}$, thus the presence of the joint layer is implicit.
The optimal depth of the \mtl heads, in our scenario, is $0$, which means that only the output layers are disjointed.
The optimal value of the width of the prediction heads is 32 neurons, whereas the width of the backbone is 64 neurons.
In the \tl case, the depth of the network is the same as the \resnet discussed in Section~\ref{subsec:results-resnet}, for compatibility between the architectures.
Moreover, the best results of \tl have been found when freezing all the layers except the output one.

\mtl shows a slight improvement in the performance on the \chf prediction obtained with the single-task network, and \tl leads to another small decrease of the \rmspe, despite using a shallower network.
However, it presents much wider uncertainty bounds than the other methods with much smaller variability.
Finally, to larger values of \picp, found in the \tl case, correspond wider bounds, as we could expect.
Although a significant difference between \tl and the other two methods can be seen, the coverage is greater than the target threshold in every case, resulting in an underconfident estimation of the uncertainty.

\begin{figure}[t]
    \centering
    \includegraphics[width=0.6\linewidth]{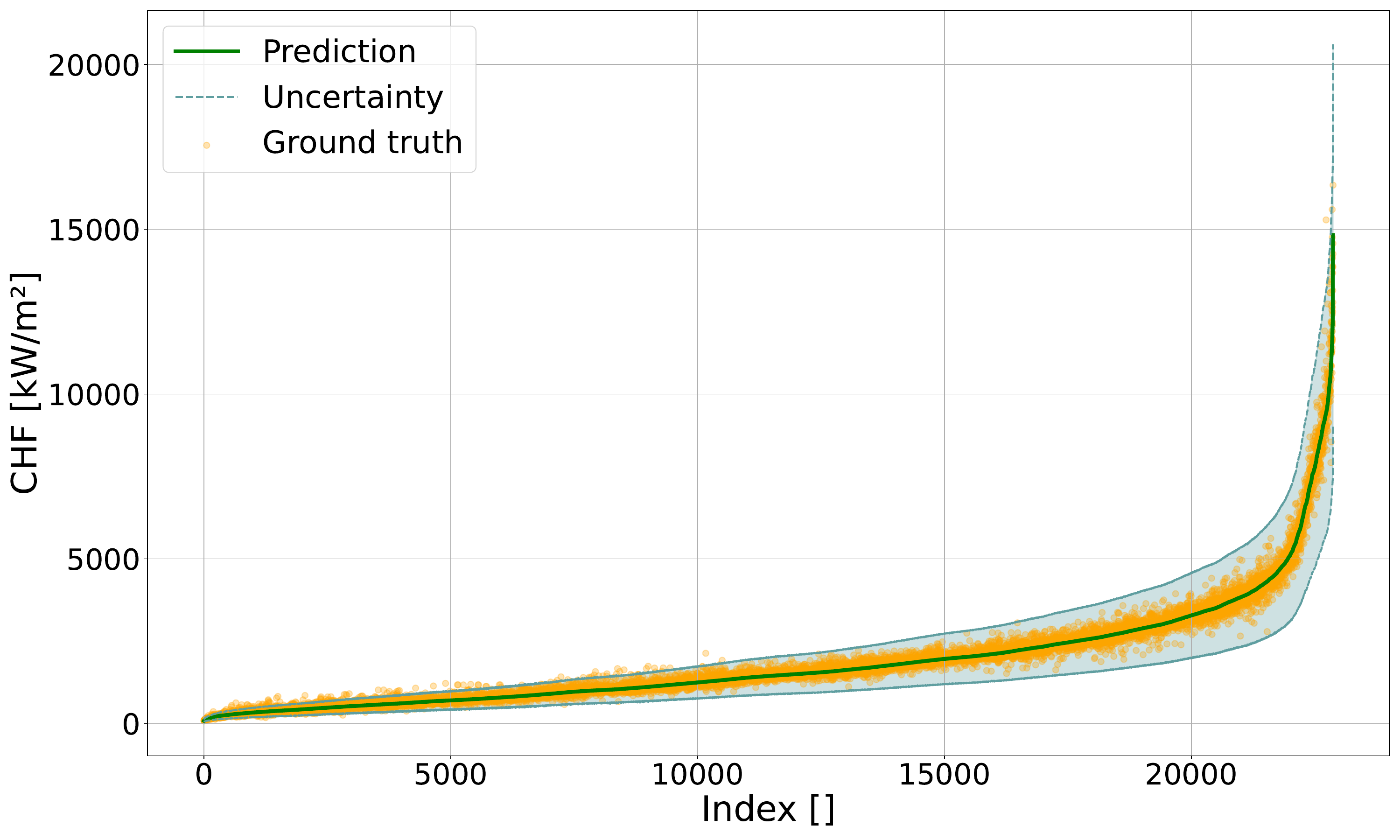}
    \caption{Uncertainty bounds of \hr with \tl.}\label{fig:enveloping-hr-transfer-learning}
\end{figure}

Figure~\ref{fig:enveloping-hr-transfer-learning} presents the results of the \resnet used in \hr.
As for Figure~\ref{fig:enveloping-cp-adaptive}, we show the predictions, the uncertainty bounds and the ground truths on the same plot.
Contrary to what was previously presented in the adaptive \cpred case, we can observe that the variability of the bounds clearly depends on the predicted \chf value: they become larger as the \chf increases.
Indeed, the few uncovered experimental points correspond to low values of predicted \chf, whereas the larger ones are mostly inside the hull.
However, such a result suggests that the \uq does not have homogeneous behaviour in this case, since it seems that the width of the bounds is underestimated for small \chf values and overestimated for large ones.

\begin{figure}[t]
    \centering
    \includegraphics[width=0.8\linewidth]{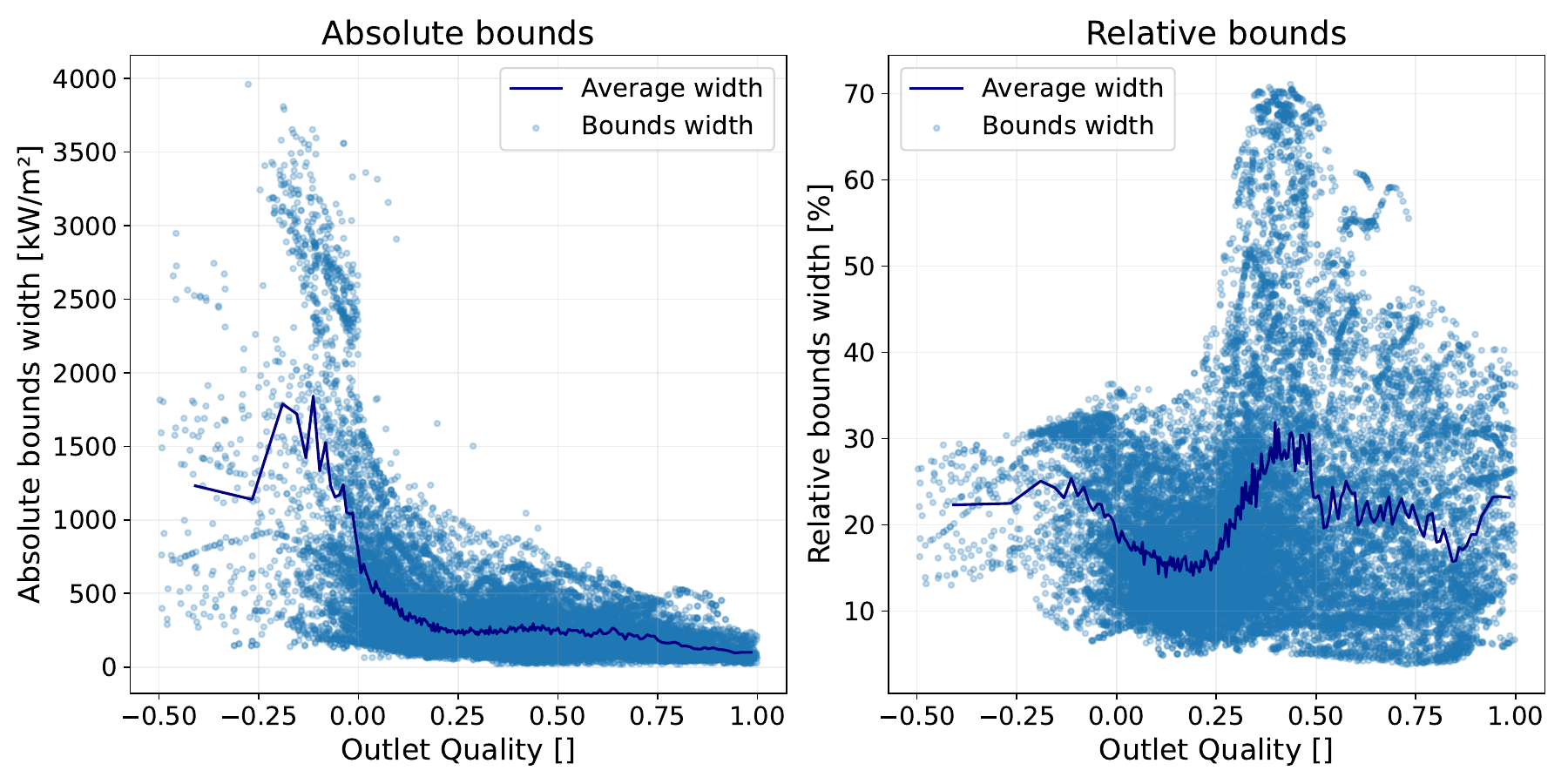}
    \caption{Uncertainty bounds vs outlet quality for \hr trained from scratch.}\label{fig:hr-whole-X-complete}
\end{figure}

As for adaptive \cpred, only the output quality $\mathfrak{X}$ shows to be connected with the relative width of the uncertainty bounds.
Figure~\ref{fig:hr-whole-X-complete} (right plot, relative bounds width), which presents the trend of the uncertainty bounds with respect to $\mathfrak{X}$, allows us to define three regions belonging to the domain of the outlet quality:

\begin{enumerate}
    \item for $\mathfrak{X} < 0.25$, the relative amplitude is usually lower, approximately delimited between 10\% and 30\%, and with very few or no outliers,
    \item for the \emph{transition regime} ($\mathfrak{X} \in [0.25, 0.50]$), the uncertainty bounds are larger, with a variable peak depending on case-by-case basis,
    \item for $\mathfrak{X} > 0.50$, the amplitude of the bounds is moderate, with generally more scattered points than in the first interval.
\end{enumerate}

As in the previous case, these results are consistent with the physical interpretation of the \chf, where the transition regime between \dnb and \dout is characterised by higher uncertainty.
This shows that the change in the internal data representation of the model, due to the joint training of the \chf prediction and the \uq, allows us to capture the complex physical behaviour of the phenomenon, which is reflected in the uncertainty bounds.
In the particular case of \hr, this is achieved without needing a post-hoc calibration step.

As observed for the standalone \resnet, discussed in Section~\ref{subsec:results-resnet}, these models are well-calibrated, since they present an \auce below \num{0.005} on the test set.
Table~\ref{tab:hr-inf-cal} shows that \tl leads to weaker informativeness, according to its wider bounds, but larger calibration compared to the other strategies, due to its better performance on \chf.
By consequence, the \uqf score is higher in the vanilla and \mtl cases, highlighting the trade-off between informativeness and calibration.
\tl presents however a much higher calibration than the other methods and wider bounds.

\begin{table}[t]
    \centering
    \caption{\uq metrics for \hr.}\label{tab:hr-inf-cal}
    \begin{tabular}{@{}ccccc@{}}
        \toprule
        \textbf{Variant} & \textbf{\rmspe} (\%) & \textbf{\inform} (\%) & \textbf{\clb} (\%) & \textbf{\uqf} (\%) \\
        \midrule
        Base             & 10.8                 & 79.2                  & 73.7               & \textbf{74.1}      \\
        \mtl             & 10.6                 & 80.3                  & 72.5               & 73.8               \\
        \tl              & \textbf{10.3}        & 60.4                  & 85.5               & 69.9               \\
        \bottomrule
    \end{tabular}
\end{table}

To summarise, \hr presents a performance close to the standalone \resnet, but it jointly performs the \chf prediction task, and the \uq.
Compared to \cpred, it provides similar or higher coverage.
It generally shows the same trend with respect to the outlet quality.
\tl is beneficial for the quality of the predictions.
Models are well-calibrated, with an \auce comparable to that found for the standalone \resnet.
Finally, the \uqf scores found for \hr are higher than those achieved with a vanilla approach to \cpred, while those for the vanilla and \mtl \hr versions are even higher than the adaptive \cpred.
Though, it must be noted that the \inform values vary with respect to the \cpred variants.
This is a standard trade-off observed in \uq tasks when balancing good informativeness and calibration, on the same plane as the precision-recall trade-off usually witnessed in classification tasks.
In a scenario where nuclear safety is involved, it is reasonable to prioritise calibration over informativeness, to avoid risky underestimations of the uncertainty.
In general, the data representations learnt by the \mtl and \tl architectures are more informative and capable of better balancing the trade-off between calibration and informativeness, compared to the vanilla \hr case.
In real-world applications, where the training data may be limited, or training costs need to be reduced, \tl delivers good results with a shallower architecture.
Conversely, the more expensive \mtl approach might be more beneficial in controlled environments, where the informativeness of the uncertainty bounds is more relevant.
It also offers wider flexibility in the design of the architecture, which can be adapted and modified for other downstream tasks, without the need to retrain the whole model from scratch.

\subsection{Quality-Driven Prediction and Uncertainty Estimation}\label{subsec:results-qd}

The application of the \qd approach is performed analogously to \hr, by investigating a baseline case in which the three outputs of the model are directly generated by the output layer after training from scratch, a variant using \mtl with one prediction head to estimate the \chf value and one to estimate the uncertainty bounds, and a variant employing \tl on the single-task architecture.
The early stopping technique is implemented with a patience of \num{100} epochs.
Hyperparameter optimisation is performed as described in Section~\ref{subsec:results-hr}.
The weight factor $\gamma$ mitigates the different scales of the loss components: since the upper bound of the penalty term in $L_{QD}$ in~\eqref{eq:qd-loss} can be expressed as $\lambda\, n \left(\frac{1}{\alpha} - 1\right) \approx 10^4\, \lambda $, then we should consider $\gamma$ such that $\left(1 - \gamma\right)\, \lambda \approx 10^{-4}$.

Thus, hyperparameter tuning involves:

\begin{enumerate}
    \item the coverage penalty factor $\lambda$, in the range $\left[ 10^1, 10^4 \right]$,
    \item the sigmoid smoothness factor $s$, in a potentially unlimited range (i.e.\ a very large range of values), to find the best possible approximation of $\vb{k}$ in~\eqref{eq:qd-loss-k-soft}, without incurring in the vanishing gradient phenomenon,
    \item the weight factor $\gamma$, in a range such that $\qty(1 - \gamma)\, \lambda \in \left[ 10^{-5}, 10^{-2} \right]$.
\end{enumerate}

Model architecture is as in Section~\ref{subsec:results-resnet}.
The width and depth of the network, learning rate, and weight decay are subject to fewer tuning procedures, starting from the configurations found in the previous numerical experiments, which show better results than others.

\begin{table}[t]
    \centering
    \caption{Hyperparameters for \qd uncertainty estimation.}\label{tab:qd-hp}
    \begin{tabular}{@{}ccccc@{}}
        \toprule
        \textbf{Variant} & \textbf{Depth} & $\lambda$ & $s$        & $1 - \gamma$ \\
        \midrule
        Base             & 8              & \num{1e2} & \num{1e-1} & \num{1e-6}   \\
        \mtl             & 6 + 1          & \num{1e3} & \num{1e-1} & \num{1e-6}   \\
        \tl              & 8              & \num{1e3} & \num{1e-1} & \num{1e-6}   \\
        \bottomrule
    \end{tabular}
\end{table}

Table~\ref{tab:qd-hp} shows the optimal hyperparameters for the three investigated variants of the \qd approach.
The depth of the network is globally equivalent across the variants, since the joint layer is implicitly counted in the \mtl architecture, for which the depth is expressed as described in Section~\ref{subsec:results-hr}.
The smoothing factor of the sigmoid $s$ and the scaling factor $1 - \gamma$ do not change for the different strategies, while the coverage penalty $\lambda$ is smaller in the baseline case.
These models are optimised using Adam, with \num{1e3} as learning rate and \num{1e-7} as weight decay.
The width of each layer is 64, including the prediction heads in the \mtl case.
When implementing \tl, the best results are found by leaving all the layers of the model unfrozen.

\begin{table}[t]
    \centering
    \caption{Results of \qd uncertainty estimation.}\label{tab:qd-results}
    \begin{tabular}{@{}cccccccc@{}}
        \toprule
        \multirow{3}{*}{\textbf{Variant}}
             & \multirow{2}{*}{\textbf{\rmspe}}
             & \multicolumn{4}{c}{\textbf{Bounds} (\%)}
             & \multicolumn{2}{c}{\textbf{Coverage} (\%)}                                                                                                                                                                  \\
        \cmidrule(ll){3-4} \cmidrule(ll){5-6} \cmidrule(ll){7-8}
             & \multirow{2}{*}{(\%)}                      & \multicolumn{2}{c}{\textbf{Lower}} & \multicolumn{2}{c}{\textbf{Upper}} & \multirow{2}{*}{\textbf{Test set}} & \multirow{2}{*}{\textbf{Dataset}}               \\
        \cmidrule(ll){3-4} \cmidrule(ll){5-6}
             &                                            & \textbf{Mean}                      & \textbf{Std}                       & \textbf{Mean}                      & \textbf{Std}                      &      &      \\
        \midrule
        Base & \textbf{10.1}                              & 19.2                               & 6.3                                & 28.7                               & 12.8                              & 96.0 & 97.8 \\
        \mtl & 10.5                                       & 28.7                               & 11.8                               & 38.2                               & 21.9                              & 96.2 & 96.9 \\
        \tl  & 10.2                                       & 21.6                               & 5.1                                & 35.0                               & 16.5                              & 97.6 & 98.9 \\
        \bottomrule
    \end{tabular}
\end{table}

\begin{figure}[t]
    \centering
    \includegraphics[width=0.6\linewidth]{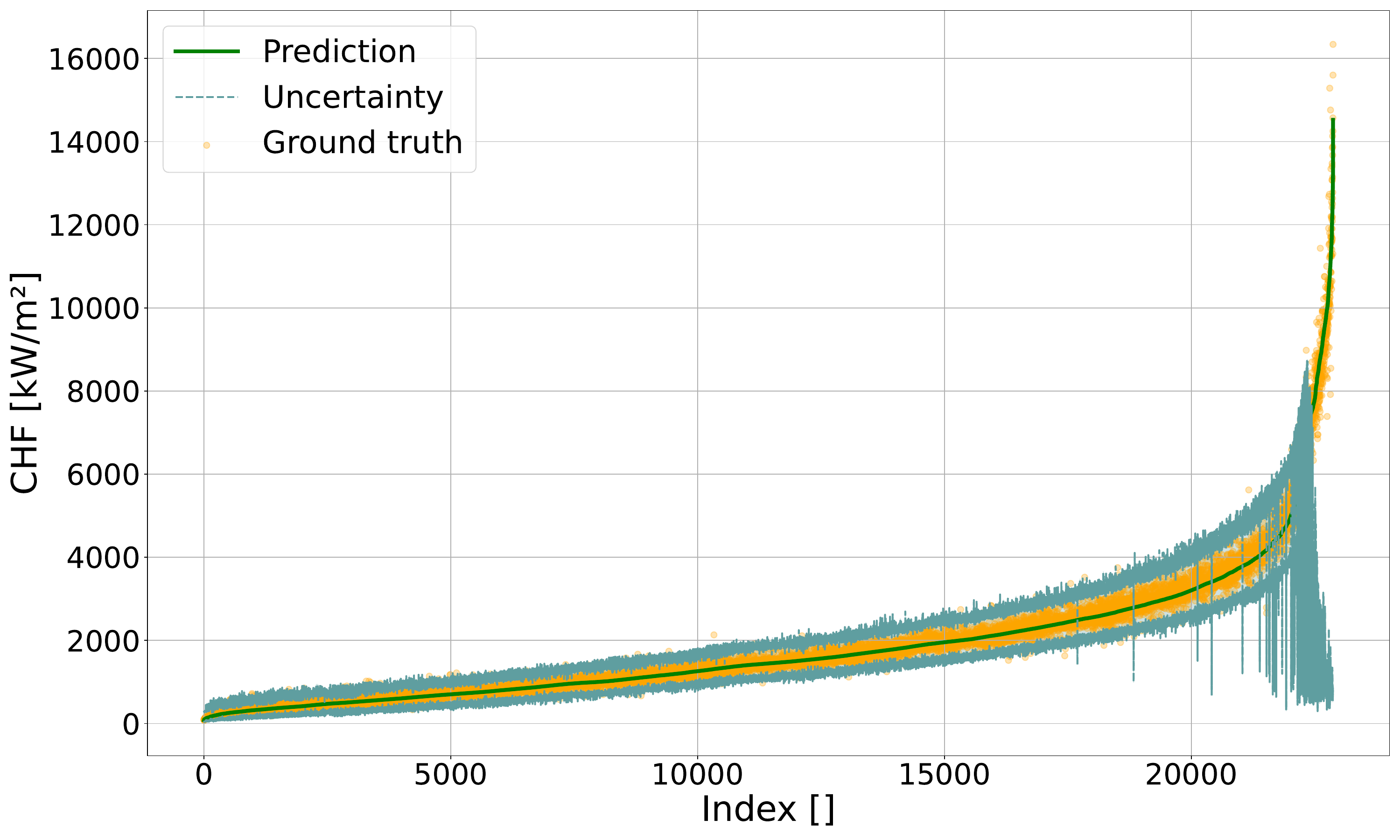}
    \caption{Uncertainty bounds of \qd uncertainty estimation with \mtl.}\label{fig:resnet-double-enveloping-qd-scratch}
\end{figure}

Table~\ref{tab:qd-results} shows the results of the three \qd variants discussed in this section, with the single-task architecture achieving the same performance as the standalone \resnet, presented in Section~\ref{subsec:results-resnet}.
Different from \hr, \mtl is slightly harmful to the network.
\tl still does not improve the single-task performance, but it achieves comparable performance.
The \mtl network presents wider uncertainty bounds, despite achieving similar coverage to the single-head architecture, which might suggest an improved adjustment of the uncertainty bounds, according to the different physical regimes, as shown in the rest of the section.
Moreover, Figure~\ref{fig:resnet-double-enveloping-qd-scratch} shows that the model is unable to estimate the uncertainty bounds for large \chf values, where the heads of the network struggle to maintain a coherent behaviour due to the strongly different scales of their predictions.
This shows that the \qd approach is more sensitive to the choice of the hyperparameters, and to the training procedure, than the other methods.
In this case, we thus flag as ``unreliable'' the predictions for which the predicted upper bound is smaller than the predicted \chf value (598 cases, corresponding to 2.6\% of the samples).
The application of \tl leads to slightly higher coverage than the single-task network, with larger uncertainty bounds.
Upper bounds are generally wider than lower ones, probably due to the fact that \chf was forced to be always positive in the network, thus more inclined to be overestimated rather than underestimated.

\begin{figure}[t]
    \centering
    \includegraphics[width=0.8\linewidth]{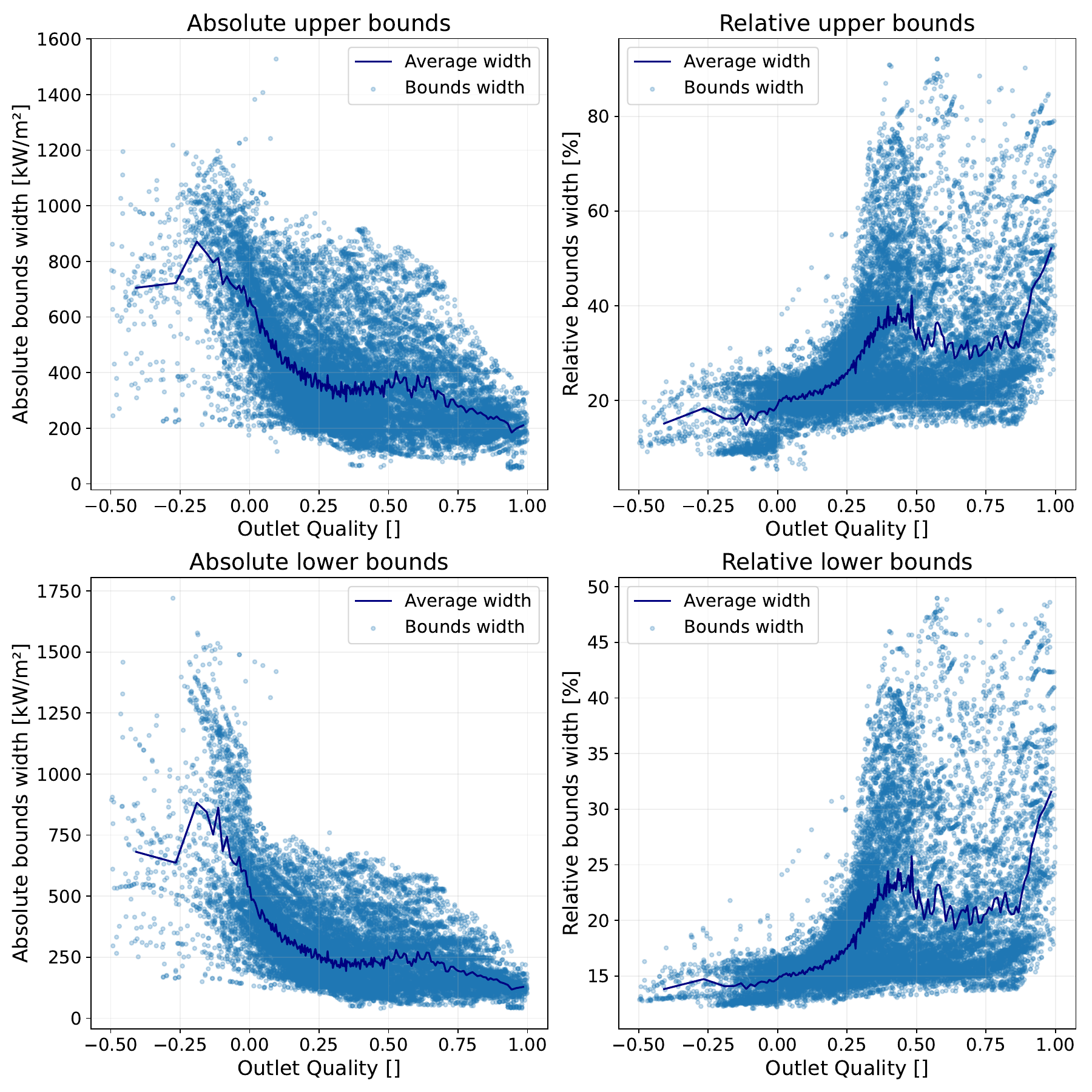}
    \caption{Uncertainty bounds vs outlet quality for \qd uncertainty estimation trained from scratch.}\label{fig:qd-scratch-whole-X-complete}
\end{figure}

As for previous \uq techniques, only the outlet quality $\mathfrak{X}$ proved to be connected with the relative amplitude of the uncertainty bounds.
As we can observe on the right of Figure~\ref{fig:qd-scratch-whole-X-complete}, it is possible to divide the outlet quality domain in four regions:

\begin{enumerate}
    \item for $\mathfrak{X} < 0$, the amplitude of the bounds generally assumes the lowest values in the domain of $\mathfrak{X}$,
    \item for $\mathfrak{X} \in \left[ 0, 0.25 \right]$, the uncertainty mostly grows with $\mathfrak{X}$, with few exceptions,
    \item for the \emph{transition regime} ($\mathfrak{X} \in [0.25, 0.5]$), the width of the bounds is usually larger and subject to high variation,
    \item for $\mathfrak{X} > 0.5$, the behaviour changes depending on the training technique and on the \ml model used: in some cases the bounds tend to be narrower, even though values are subject to large variations, whereas in some others they are nearly as wide as in the transition region.
\end{enumerate}

Once again, these results are consistent with the physical interpretation of the \chf, where the transition regime between \dnb and \dout presents higher uncertainty.
As in the case of \hr, the \qd approach is able to capture this behaviour by jointly learning the \chf prediction and the uncertainty estimation, without needing a post-hoc calibration step.
However, the expressivity of the model vastly increases due to the prediction of asymetric bounds, which allows us to capture possible differences between underestimation and overestimation of the \chf predictions, as shown by the different trends of the upper and lower bounds with respect to $\mathfrak{X}$.

By considering the influence of $\mathfrak{X}$ for the baseline model, without \tl, we observe that the \dnb regions ((1) and (2)) have uncertainty values between 10\% and 30\% for the upper bounds and below 20\% for the lower ones.
The sparsity of the \dout region makes it harder to determine a threshold of $\mathfrak{X}$ for it due to its similarity with the transition interval, with peaks going above 80\% for the upper bounds and up to 50\% for the lower ones.
The average width of the bounds shows a clear trend, which starts with a small value in \dnb region, increases in intervals (2) and (3), then decreases in \dout.
A sudden growth of the average width is observed at the highest values of $\mathfrak{X}$, probably also due to the sparsity of the distribution.

Similar to the previously described \ml techniques, these models show to be well-calibrated, with an \auce value around \num{6e-3} or smaller.
As visible in Table~\ref{tab:qd-inf-cal}, informativeness drops by 5\% between the baseline approach and \tl, both based on a single-task architecture.
A strong decrease of about 4\% with respect to \tl is observed when employing \mtl.
However, this strategy showcases higher calibration, followed by \tl and then by the baseline variant.
The \uqf score, however, presents an inverted trend across the variants, thanks to the different calibration scores of \tl and the more complex \mtl architecture.

\begin{table}[t]
    \centering
    \caption{\uq metrics for \qd techniques.}\label{tab:qd-inf-cal}
    \begin{tabular}{@{}cccccc@{}}
        \toprule
        \textbf{Variant} & \textbf{\rmspe} (\%) & \textbf{\inform} (\%) & \textbf{\clb} (\%) & \textbf{\uqf} (\%) \\
        \midrule
        Base             & \textbf{10.1}        & 76.0                  & 77.1               & \textbf{73.8}      \\
        \mtl             & 10.5                 & 67.4                  & 81.2               & 70.7               \\
        \tl              & 10.2                 & 71.7                  & 78.8               & 72.9               \\
        \bottomrule
    \end{tabular}
\end{table}

To conclude the analysis of this methodology, we can affirm that the \qd approach showcases equal or slightly better performance on \chf prediction compared to the standalone \resnet model by jointly minimising the prediction error and the average amplitude of the bounds, and embedding a constraint on the coverage inside the learning process.
Moreover, this technique provides asymmetric uncertainty bounds, accounting for possible differences between underestimation and overestimation of the \chf predictions.
Lower bounds are found to be narrower than the upper ones, ranging in intervals comparable to those of \hr, and mostly narrower than \cpred, apart from specific cases (e.g.\ \mtl in the \qd approach).
\mtl and \tl finally do not improve the performance on \chf, but they provide a better calibrated estimation of the uncertainty, thanks to the more informative data representations learnt by the network.

\subsection{Bayesian Heteroscedastic Regression}\label{subsec:results-bhr}

\bhr is implemented analogously to \hr, by using a \resnet-based \bnn with pre-activated residual blocks and batch normalisation.
In this section, both single-task and \mtl architectures are investigated, and their uncertainty components are compared (\tl cannot be applied in this case).
The training process is executed with a patience of 100 epochs and using a batch size of 1024 samples.
Hyperparameter tuning is performed by computing the \rmspe and the coverage for each epoch on the validation set.
The model that achieves the best performance on \chf while reaching at least the target coverage of 95\% is selected.
Hyperparameter tuning mainly involves the learning rate in $\left[ 10^{-4}, 10^{-3}\right]$, the weight decay in the space $\left[ 0, 10^{-6}\right]$, the regularization factor of the \kl divergence $\beta_{\text{\kl}}$ in the range $\left[10^{-3}, 10\right]$, the width of the network in $\left[32, 64\right]$, the depth of the network in $\left[ 8, 20 \right]$ for the single task architecture, and, in the \mtl case, the depth of the backbone and of the prediction heads in the ranges $\left[7, 11\right]$ and $\left[0, 2\right]$ respectively.
The models are optimised with Adam, with an optimal learning rate of \num{1e-3} and a weight decay of \num{1e-7}.
The optimal width of the network is 64 neurons per layer, including that of the prediction heads in the \mtl model.
The chosen values of the other hyperparameters are reported in Table~\ref{tab:bhr-results}, together with the results.
The information on the depth of the network is presented as described in Sections~\ref{subsec:results-hr}~and~\ref{subsec:results-qd}.

\begin{table}[t]
    \centering
    \caption{Hyperparameters and results for \bhr.}\label{tab:bhr-results}
    \begin{tabular}{@{}cccccccc@{}}
        \toprule
        \multirow{2}{*}{\textbf{Variant}}
             & \multirow{2}{*}{\textbf{Depth}}
             & \multirow{2}{*}{$\boldsymbol{\beta}_{\text{\kl}}$}
             & \multirow{2}{*}{\textbf{\rmspe} (\%)}
             & \multicolumn{2}{c}{\textbf{Bounds} (\%)}
             & \multicolumn{2}{c}{\textbf{Coverage} (\%)}                                                                                                            \\
        \cmidrule(ll){5-8}
             &                                                    &            &               & \textbf{Mean} & \textbf{Std} & \textbf{Test set} & \textbf{Dataset} \\
        \midrule
        Base & 12                                                 & 1          & \textbf{11.1} & 17.5          & 9.4          & 95.6              & 97.9             \\
        \mtl & 11 + 0                                             & \num{1e-3} & 11.3          & 18.7          & 9.7          & 95.4              & 97.8             \\
        \bottomrule
    \end{tabular}
\end{table}

\begin{figure}[t]
    \centering
    \includegraphics[width=0.7\linewidth]{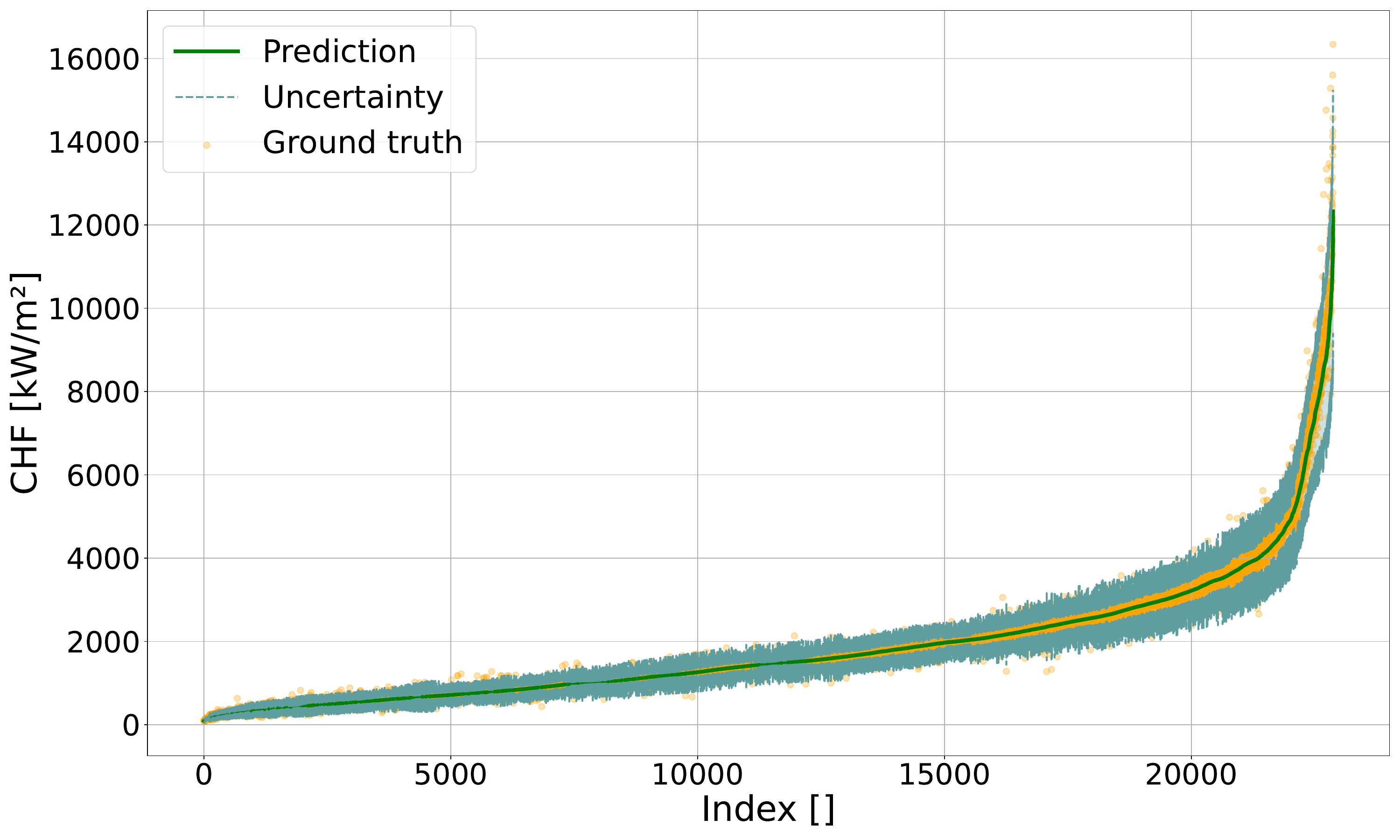}
    \caption{Uncertainty bounds of \bhr, with single-head architecture.}\label{fig:bhr-enveloping-single}
\end{figure}

We can observe that single-head architecture performs slightly worse than the \resnet-based \hr discussed in Section~\ref{subsec:results-hr}, though the target coverage is always reached.
Contrary to \hr, the \mtl approach does not improve the performance on \chf.
For both types of architectures, the overall shape of the uncertainty bounds are similar to the corresponding non-Bayesian approach (the convex hull of the bounds shown in Figure~\ref{fig:enveloping-hr-transfer-learning} and Figure~\ref{fig:bhr-enveloping-single}).

\begin{table}[t]
    \centering
    \caption{Contribution of uncertainty components in \bhr.}\label{tab:bhr-uq}
    \begin{tabular}{@{}ccccccc@{}}
        \toprule
        \multirow{2}{*}{\textbf{Variant}}
             & \multicolumn{3}{c}{\textbf{Test set} (\%)}
             & \multicolumn{3}{c}{\textbf{Dataset} (\%)}                                                                                                    \\
        \cmidrule(ll){2-7}
             & \textbf{Prediction}                        & \textbf{Model} & \textbf{Aleatoric} & \textbf{Prediction} & \textbf{Model} & \textbf{Aleatoric} \\
        \midrule
        Base & 8.9                                        & 2.2            & 8.6                & 8.9                 & 2.2            & 8.6                \\
        \mtl & 9.6                                        & 2.4            & 9.2                & 9.7                 & 2.4            & 9.2                \\
        \bottomrule
    \end{tabular}
\end{table}

Table~\ref{tab:bhr-uq} shows the contribution of the two components of uncertainty (\emph{model} and \emph{aleatoric}) for both the \bhr models, and the corresponding prediction uncertainty, used to compute the bounds as described in Section~\ref{subsec:method-bhr}.
As we could have expected by observing the average bounds width in Table~\ref{tab:bhr-results}, the double-head model has larger uncertainty components than the single-head one.
In both cases, aleatoric uncertainty contributes much more to prediction uncertainty than the model component.
This result is consistent with both the generally small model uncertainty found by previous works~\cite{furlong-pinn-uq, alsafadi-cvae}, and the well-known unreliability of some data sources, even if the filtered dataset has been used for training (see Section~\ref{sec:dataset}).

\begin{figure}[t]
    \centering
    \includegraphics[width=0.8\linewidth]{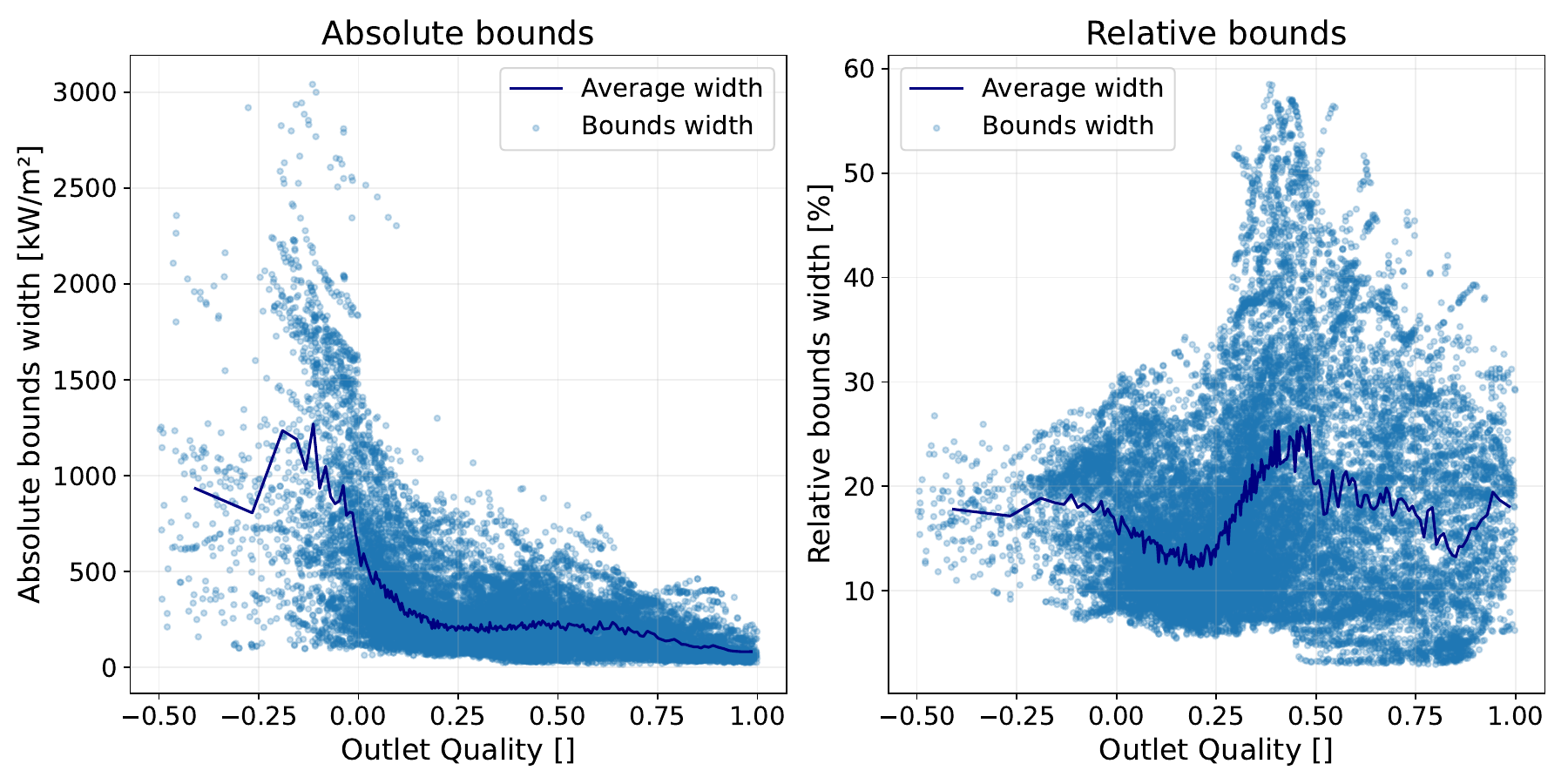}
    \caption{Uncertainty bounds vs outlet quality for \bhr with single-head architecture.}\label{fig:bhr-single-whole-X}
\end{figure}

Figure~\ref{fig:bhr-enveloping-single} shows the uncertainty bounds around the predictions for the best performing model, that is, the single-head \bnn.
We can notice the variability of the bounds, whose width generally increases with the \chf prediction.
However, no strict dependence can be observed between these quantities, unlike Figure~\ref{fig:enveloping-hr-transfer-learning}.
Ground truth points (in yellow) are mostly enveloped by the bounds, according to the high coverage we found (see Table~\ref{tab:bhr-results}).
As in the previous cases, sensitivity analysis revealed that only the outlet quality $\mathfrak{X}$ presents a significant connection with the relative width of the uncertainty bounds.
Figure~\ref{fig:bhr-single-whole-X} (right plot, relative bounds width) enables to divide the outlet quality domain in three regimes:

\begin{enumerate}
    \item for $\mathfrak{X} < 0.25$, the relative width is usually small, approximately delimited between the 5\% and the 30\%, with very few or no outliers, and whose maximum values increase with $\mathfrak{X}$,
    \item for the \emph{transition regime} ($\mathfrak{X} \in [0.25, 0.50]$), the uncertainty bounds are larger, with values reaching 60\%,
    \item for $\mathfrak{X} > 0.50 $, the amplitude is generally moderate, with more sparse points than in the first interval.
\end{enumerate}

As in the previous cases, these results are consistent with the physical interpretation of \dnb and \dout regimes.
This shows again that the change in the internal data representation of the model, due to the joint training of the \chf prediction and the \uq, allows us to capture the complex physical behaviour of the phenomenon, which is reflected in the uncertainty bounds.

The average value of the width shows the same trend observed in both adaptive \cpred (see Section~\ref{subsec:results-cp}) and in \hr (see Section~\ref{subsec:results-hr}): the average width of the bounds is almost constant in the \dnb region, then grows before decreasing again in the \dout region.
Finally, a sudden growth is observed for the highest outlet quality values.
The \mtl architecture, which is not analysed in detail here, provides a similar trend.
The relative amplitude of the bounds is smaller than the adaptive \cpred (Figure~\ref{fig:cp-whole-X}) and than \hr (Figure~\ref{fig:hr-whole-X-complete}).
Both \bhr approaches show to provide well-calibrated predictions, as previously described methodologies.
In this case, the \auce remains below \num{0.005}.

\begin{table}[t]
    \centering
    \caption{\uq metrics of \bhr.}\label{tab:bhr-inf-cal}
    \begin{tabular}{@{}ccccc@{}}
        \toprule
        \textbf{Variant} & \textbf{\rmspe} (\%) & \textbf{\inform} (\%) & \textbf{\clb} (\%) & \textbf{\uqf} (\%) \\
        \midrule
        Base             & \textbf{11.1}        & 82.5                  & 69.4               & \textbf{72.5}      \\
        \mtl             & 11.3                 & 81.3                  & 67.4               & 70.6               \\
        \bottomrule
    \end{tabular}
\end{table}

For each of the two \bhr variants, Table~\ref{tab:bhr-inf-cal} presents the relative informativeness, the uncertainty calibration and the \uqf score on the whole dataset.
The informativeness is not significantly influenced by the architecture used (1\% drop at most when using \mtl), whereas the calibration (and thus the \uqf score) decreases when employing the double head.
The results observed for these models vary across different variants, in terms of \uqf: the base version performs worse than the base \qd models and the vanilla \hr, while performing better than the \cpred variants.
On the other hand, the \mtl version is worse than the \mtl version of \hr and adaptive \cpred, and equivalent to the \mtl \qd model.

Thus, \bhr provides a framework to perform joint regression and \uq, as potential alternative to \hr and the \qd approach, whose results are discussed in Sections~\ref{subsec:results-hr} and~\ref{subsec:results-qd}, respectively.
The main feature added by \bhr is the possibility of separately capturing aleatoric (data) and epistemic (model) uncertainty, which can be used to evaluate the convergence of the model and the trustworthiness of its predictions.
This is possible due to the probabilistic assumptions on which \bnn architectures are based.
This can cause the optimisation process to be slow and non-trivial.
However, \bhr showed nearly the same results as \hr on the \chf prediction task, providing the narrowest uncertainty bounds despite reaching the target coverage.
Moreover, \bhr offered the possibility of quantitatively estimating the model contribution to the uncertainty, which came out to be much smaller than the aleatoric component, as expected.

\section{Conclusion}\label{sec:conclusion}

This work addressed two fundamental challenges in scientific \ml: the accurate prediction of complex physical phenomena and the trustworthy quantification of their uncertainty.
We employed the prediction of \chf via the \nrc dataset as a benchmark to evaluate our general approach to multi-regime physical systems.
By carefully tidying the dataset and restoring physical consistency to the training domain, specifically through the removal of pathological entries and non-physical negative subcooling data, we enabled our \resnet models, explored across various architectural configurations, to achieve a predictive error (\rmspe) of approximately 10\%.
This performance aligns with, and in some configurations exceeds, the current state-of-the-art, confirming that the baseline predictive power of the architecture is maximised for the downstream task of predicting the value of the \chf.

Beyond global accuracy, our comparative analysis of \uq methodologies reveals a critical distinction between post-hoc calibration and coverage-oriented learning.
While adaptive \cpred successfully ensures statistical safety by calibrating a frozen model, end-to-end approaches such as \hr and \qd learning force the model to internalise the physical profile of the data.
This transition to heteroscedastic modelling was accompanied by rigorous statistical testing of the data behaviour, which confirmed the non-uniform variance across the input space.
The uncertainty bounds found by our models were found to be highly consistent with this statistical signature, allowing them to autonomously identify the physical transition between the stable \dnb regime and the onset of \dout.
This discovery manifested as a distinct, localised increase in aleatoric uncertainty in the transition zone (outlet quality $\mathfrak{X} \in [0.25, 0.5]$), effectively flagging the regime change without explicit supervision.
Furthermore, our Bayesian analysis (\bhr) confirmed that the total uncertainty is dominated by the aleatoric component, validating the hypothesis that the variability is intrinsic to the physics rather than a product of model ignorance.

These findings suggest a shift in how \uq could be viewed in scientific and engineering applications.
It is not merely a final safety check, that is a post-hoc (``plug-in'') technique to satisfy the \bepu framework requirement, but an active component of the learning process that transforms the model into a self-diagnosing tool capable of revealing physical regimes and associated transitions.
By bridging the gap between black-box accuracy and phenomenological consistency, coverage-oriented \uq offers a robust framework for learning reliable physical behaviours from data, particularly in high-stakes environments (such as nuclear engineering) where understanding the boundaries of knowledge is as important as the prediction itself.

Ultimately, this study underscores the pivotal role of representation learning in capturing the heterogeneous nature of complex physical systems.
Our results demonstrate that integrating uncertainty quantification directly into the learning objective is not only a path to safer predictions but also a powerful strategy to enrich the model's internal representation.
By forcing the architecture to account for the varying physical regimes of the data, we enable it to learn a more complete and statistically robust mapping of the observed phenomenon.
This capability to highlight and learn data-driven behaviours could ultimately lead to more interpretable and reliable scientific machine learning models.

\section*{CRediT Author Statement}
\textbf{Michele Cazzola}: Conceptualization, Methodology, Software, Validation, Formal analysis, Investigation, Data curation, Writing --- original draft, Writing --- Review \& Editing, Visualization. \textbf{Alberto Ghione}: Conceptualization, Methodology, Validation, Formal analysis, Investigation, Data curation, Writing --- Review \& Editing, Supervision. \textbf{Lucia Sargentini}: Conceptualization, Methodology, Validation, Formal analysis, Investigation, Data curation, Writing --- Review \& Editing, Supervision. \textbf{Julien Nespoulous}: Conceptualization, Methodology, Validation, Formal analysis, Investigation, Data curation, Writing --- Review \& Editing, Supervision. \textbf{Riccardo Finotello}: Conceptualization, Methodology, Validation, Formal analysis, Investigation, Data curation, Writing --- Original draft, Writing --- Review \& Editing, Visualization, Supervision.

\section*{References}
\printbibliography[heading=none]

\end{document}